\documentclass[sigconf]{acmart}
\usepackage{multirow}
\usepackage{graphicx}
\usepackage{enumitem}
\AtBeginDocument{%
  }

\copyrightyear{2025}
\acmYear{2025}
\setcopyright{cc}
\setcctype{by-nd}
\acmConference[WWW '25]{Proceedings of the ACM Web Conference 2025}{April 28-May 2, 2025}{Sydney, NSW, Australia}
\acmBooktitle{Proceedings of the ACM Web Conference 2025 (WWW '25), April 28-May 2, 2025, Sydney, NSW, Australia}
\acmDOI{10.1145/3696410.3714528}
\acmISBN{979-8-4007-1274-6/25/04}

\begin{document}


\title{Unmasking Gender Bias in Recommendation Systems and Enhancing Category-Aware Fairness}

\author{Tahsin Alamgir Kheya}
\email{t.kheya@deakin.edu.au}
\orcid{0009-0001-2481-2877}
\affiliation{%
  \institution{Deakin University}
  \city{Geelong}
  \state{VIC}
  \country{Australia}
}

\author{Mohamed Reda Bouadjenek}
\email{reda.bouadjenek@deakin.edu.au}
\orcid{0000-0003-1807-430X}
\affiliation{%
  \institution{Deakin University}
  \city{Geelong}
  \state{VIC}
  \country{Australia}}

\author{Sunil Aryal}
\email{sunil.aryal@deakin.edu.au}
\orcid{0000-0002-6639-6824}
\affiliation{%
  \institution{Deakin University}
  \city{Geelong}
  \state{VIC}
  \country{Australia}
}

\renewcommand{\shortauthors}{Tahsin Alamgir Kheya, Mohamed Reda Bouadjenek, and Sunil Aryal}


\begin{abstract}
Recommendation systems are now an integral part of our daily lives. We rely on them for tasks such as discovering new movies, finding friends on social media, and connecting job seekers with relevant opportunities.
Given their vital role, we must ensure these recommendations are free from societal stereotypes. 
Therefore, evaluating and addressing such biases in recommendation systems is crucial. 
Previous work evaluating the fairness of recommended items fails to capture certain nuances as they mainly focus on comparing performance metrics for different sensitive groups.
In this paper, we introduce a set of comprehensive metrics for quantifying gender bias in recommendations.
Specifically, we show the importance of evaluating fairness on a more granular level, which can be achieved using our metrics to capture gender bias using categories of recommended items like genres for movies. Furthermore, we show that employing a category-aware fairness metric as a regularization term along with the main recommendation loss during training can help effectively minimize bias in the models' output.
We experiment on three real-world datasets, using five baseline models alongside two popular fairness-aware models, to show the effectiveness of our metrics in evaluating gender bias. 
Our metrics help provide an enhanced insight into bias in recommended items compared to previous metrics. Additionally, our results demonstrate how incorporating our regularization term significantly improves the fairness in recommendations for different categories without substantial degradation in overall recommendation performance.
\end{abstract}




 \maketitle
 \textit{This paper has been accepted for the ACM Web Conference 2025.}

\section{Introduction}
Recommender Systems (RS) personalize item selections for users, providing suggestions based on individual preferences and behaviors. 
These systems have an important impact on our decision-making, as they are widely employed across diverse platforms like e-commerce, social media, streaming services, and news outlets, shaping the content and products we encounter.
For several online platforms, recommendation systems help create an engaging experience for users by diversifying and personalizing the content and interactions. 
This would help users avoid information overload and help them focus on options that reflect their past behaviors. 
Amid the promise held by RS, however, there are concerns about potential bias in these systems. 
For example, research has shown that, on certain recommender systems, simply changing the gender in a job search—while keeping qualifications constant—can significantly influence access to high-paying positions \cite{datta15,10.1145/3442381.3450077}.

\begin{figure}[t]
    \centering
    \includegraphics[width=0.5\textwidth]{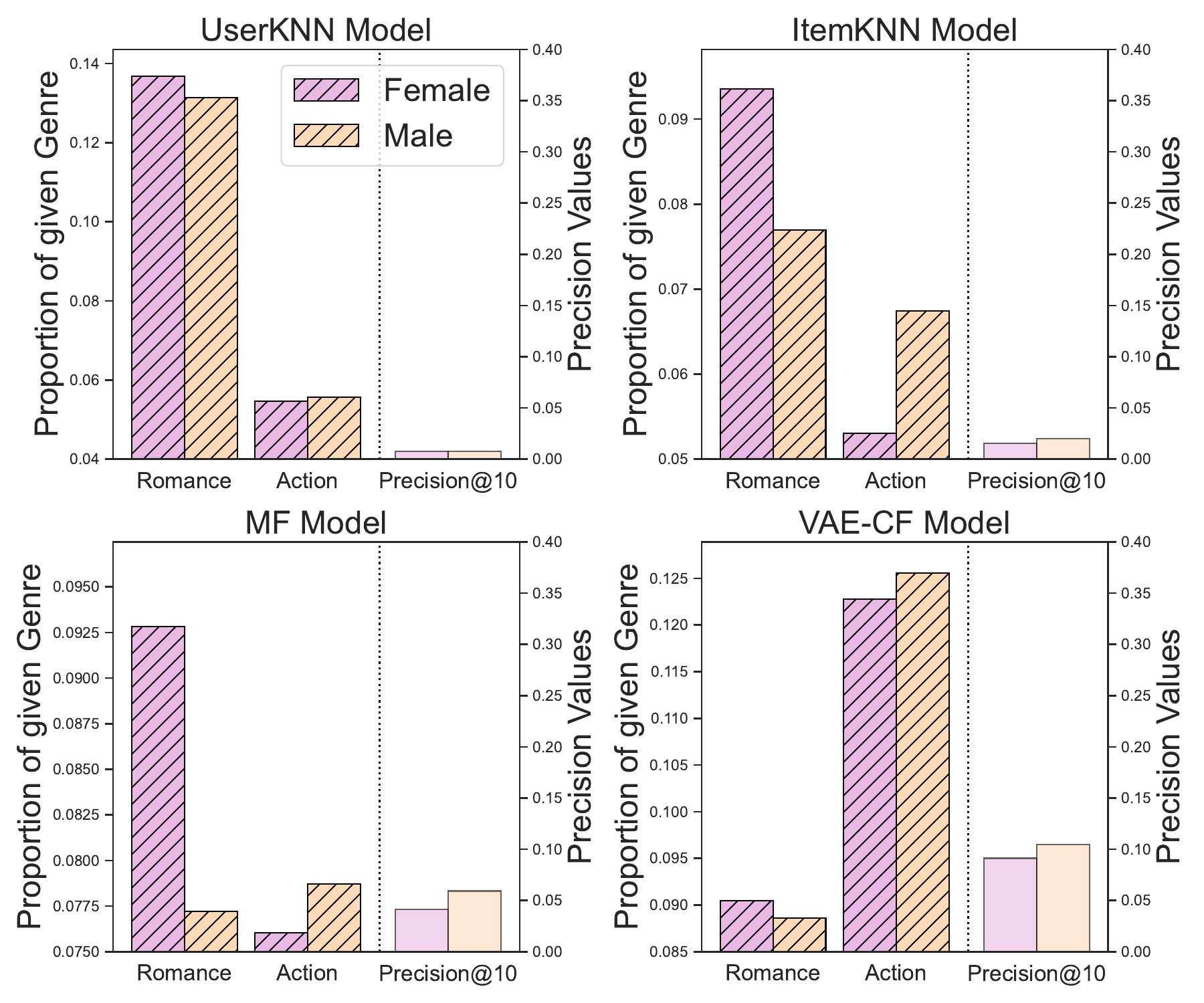}
    \caption{
    Comparison of action and romance movie recommendations among male and female users across four recommendation algorithms, along with corresponding Precision@10 values. The graphs highlight disparities in genre recommendations by gender, with romance movies being more frequently suggested to female users and action movies to male users, despite similar Precision@10 metrics. }
    \label{fig:compare_genre}
\end{figure}

RS algorithms are traditionally evaluated using measures like RMSE (Root Mean Squared Error), NDCG (Normalized Discounted Cumulative Gain), precision, and diversity \cite{Shani2021}. 
If only these metrics are considered, then the recommended items are deemed good when they align with user preferences.
These metrics can help evaluate the performance of the RS, but in recent years there is a growing emphasis on evaluating and ensuring fairness as well \cite{10.1145/3442381.3449866,10.1145/3477495.3531959,abdollahpouri2019unfairness,10.1145/3450613.3456821,NEURIPS2020_9d752cb0,10.1145/3437963.3441824}. 
For instance, the authors in \cite{deldjoo_flexible_2021} introduce a fairness evaluation metric that can be sensitive to different fairness notions like user-centric or item-centric. 
Although substantial work has been done in this field in recent years, there is a notable gap in research specifically focused on robust ways to quantify consumer bias accurately. 
Most of the metrics used are deficient in the following ways: 
(i) they over-simplify the concept of fairness,
(ii) they fail to consider the ranking of recommended items, and
(iii) they rely exclusively on a single type of fairness metric. 
A popular approach to evaluating consumer-side fairness in recommendation systems involves an adaptation of \textit{equal opportunity} (refer to Section: \ref{subsec:fair_notion}), which aims to balance performance metrics (such as recall) or utility scores across different groups, such as male vs. female users \cite{10.1145/3442381.3449866,10.1145/3655631,10.1145/3651167,10.1145/3604915.3608784,tang_when_2023,Melchiorre21}. 
However, while this approach is quite straightforward, it may overlook disparities in recommendations across different item categories.
To assess these disparities in recommendations, we refer to Figure \ref{fig:compare_genre}, where we present, for various recommendation algorithms, an analysis of the proportion of action and romance movies among the top 10 recommendations for male and female user groups, along with the corresponding Precision@10 values for each group.
There are three notable observations here:
(i) across all models, there is a noticeable disparity in the proportion of romance and action movies recommended, as romance movies tend to be recommended more frequently to female users, while action movies are more often recommended to male users;
(ii) precision@10 values for both male and female users are similar across all models, suggesting that the models appear to perform equally well for both genders based on this metric;
and (iii) different models exhibit varying levels of bias.
These observations suggest that simply comparing performance metrics across sensitive attributes is insufficient to assess fairness in recommendation systems. 
More granular metrics are necessary to accurately quantify bias and ensure fair recommendations across different user groups. To further emphasize the significance of granular evaluation, we provide a brief overview here with a more detailed example provided in Section \ref{subsec:motivating_example}. Bias can arise in recommender systems due to stereotypical interactions in the dataset used to train them, which leads to biased recommendations based on sensitive attributes of the users. This bias can trap users in filter bubbles, with a focus on stereotypical categories or narrowly defined preferences. Providing more of a certain type of content can have unintended consequences; for instance, if a young male user is predominantly recommended action movies, where violence is glorified and shown in a consequence-free way, then it could contribute to adverse effects, including the desensitization of violence. This is just one example, but the stakes are much higher for more sensitive domains like news, job recommendations, etc., where there can be catastrophic consequences if similar biases manifest in them.

In this paper, our focus is to overcome the limitations of current fairness assessment metrics by designing a set of evaluation measures to help us quantify gender bias in recommendation models. 
Specifically, our aim is to ensure that they are diverse so as to capture nuances of the bias that might not be immediately obvious but are significant.
We propose to do this by incorporating item categories and rank (when relevant). 
Next, we utilize one of these metrics as part of the loss function to optimize, which results in an effective increase of fairness for recommendations made to users for each category. 
This not only shows that our metrics are effective in addressing fairness concerns but also proves their utility for quantifying bias. Additionally, we use our metrics to evaluate a variety of recommendation models, including fairness-aware ones, using three real-world datasets. The contribution of this paper can be summarized as follows:
\begin{itemize}[left=3pt]
    \item We introduce a set of metrics that considers both categories and rankings of the recommendations made, in turn providing a more nuanced evaluation.
    \item We evaluate different recommendation algorithms to check for gender bias using the proposed set of metrics.
    \item We employ an in-processing technique to ensure the recommended items are fair, not only in a general sense but in a category-aware sense.
\end{itemize}
\section{Related work}
We categorize and discuss the related work existing in this field in the subsections below.
\subsection{Gender Fairness in RS}
Gender bias in recommendation systems can exist as systematic discrepancies in the algorithms when recommending items to users of different genders. The challenges of gender bias can have a wide-ranging set of implications, including imbalanced representations of items for different genders, stereotypical recommendations (male-dominated occupations recommended to males more than females), and limitations in user personalization (being recommended items that are not related to users' preferences just because they are male or female). Research on evaluating and addressing gender bias in recommendation systems is an ongoing process. Melchiorre et al. \cite{Melchiorre21} investigate the impact of a common de-biasing strategy called resampling on RS algorithms. This strategy marginally decreases gender bias, with a slight decrease in performance. The authors in \cite{pmlr-v139-gorantla21a,10.1145/3132847.3132938,XIA2019104857} introduce ways to re-rank items to offer a balanced solution that caters to group fairness (for gender and other sensitive attributes) and user preferences.
Historical data that can contain stereotypical movie preferences can intensify certain biases further when used to train traditional recommendation models. For instance, male users display a bias towards \textit{action} movies, which is amplified by recommendation algorithms like UserKNN \cite{TsintzouPT19}. The authors in \cite{Acharyya_Das_Chattoraj_Tanveer_2020}
introduce a framework that fairly predicts the quality of TedTalk speeches by using causal models and counterfactuals to mitigate gender and racial bias. Adversarial fairness where the recommender system is trained to not only make accurate predictions but also make it difficult for the adversary to guess the sensitive attribute, has also been employed to make such systems more fair \cite{LIU2022108058,rus2022closinggenderwagegap}. Recent advancements in this field, have led to the development of fairness-aware recommendation models, with special emphasis on gender bias \cite{yang2020causalintersectionalityfairranking, boratto2022consumer,10.1145/3397271.3401177,zhu18,wei2022comprehensive,10.1145/3442381.3450015}.
\subsection{Evaluating Gender Bias}
\label{subsec:fair_notion}
For evaluating gender bias, the most common fairness definitions employed are the concepts of \textit{Demographic Parity} and \textit{Equal Opportunity}, which are both related to group fairness. Fairness in this context, pertains to equitable treatment across different groups (which can be measured in classification and recommendation tasks).
For group fairness, the idea is to ensure that the predicted outcomes  \(\hat{Y}\) of a model should not be dependent on sensitive attributes like gender $S$. 
For demographic parity, the proportions of each sensitive group (like male and female) receiving positive predictions should be equal. 
For binary classification, demographic parity can be formalized as: 
\[P(\hat{Y}=1 | S=1) = P(\hat{Y}=1 | S=0)\] 
Essentially what this implies is that the positive outcome should be the same for both genders, where 0 may represent male and 1 may represent female or vice versa. 
Equal Opportunity, on the other hand, holds when the model has equal true positive rates across different demographic groups \cite{hardt16}. 
This concept can be formalized as:
\[P(\hat{Y}=1 | S=1, Y=1) = P(\hat{Y}=1 | S=0, Y=1) \]
where \(Y\) represents the true outcome.

Besides these two methods to quantify fairness, some additional concepts (as discussed in \cite{kheya2024pursuitfairnessartificialintelligence}) used include Equalized Odds \cite{hardt16}, Balance for Negative Class \cite{10.1145/3194770.3194776}, Balance for Positive Class \cite{10.1145/3194770.3194776}, Intersectional Fairness \cite{Gohar2023ASO}, Equal Calibration \cite{Chouldechova2016FairPW}
and Causal-based notions \cite{10.5555/3294996.3295162, Acharyya_Das_Chattoraj_Tanveer_2020}.

\subsection{Evaluating Consumer-Side Fairness}
 In recommendation systems, fairness can be seen as a multi-sided concept and categorized into three groups: consumers (C-fairness) \cite{{deldjoo_flexible_2021,Melchiorre21,wu20,pmlr-v139-gorantla21a,ghosh21,Edizel20,wan20,hao21}} which is related to the impact of recommendations of the system on protected classes of user, provider (P-fairness) which focuses on ensuring fairness for providers/sellers on a platform and both (CP-fairness) \cite{burke_multisided_2017}. Our work focuses on the consumer-side fairness concept because we want to ensure that recommendations made by models are not biased against a certain gender. Prior research has shown how recommendations can differ in an unfair way based on sensitive attributes of users like gender, age, race, etc. \cite{gupta_questioning_2022,10.1145/3547333,JIN2023101906}. Evaluating bias in these systems, before deploying is thus essential to stop the reinforcement of stereotypes and limiting diverse content. When quantifying consumer side bias in recommendation systems, the most common approach is to adopt the concept of equality of opportunity and focus on the differences in metrics such as recall, precision and/or NDCG \cite{Fu20,10.1145/3442381.3449866,Melchiorre21}. Other papers also employ causal-based fairness notions \cite{wei2022comprehensive,10.1145/3404835.3462943} and demographic parity \cite{10.1145/3292500.3330691,DBLP:journals/corr/abs-1809-09030,boratto2022consumer}. Unlike these metrics, which assume fairness implies equality, \cite{deldjoo_flexible_2021} suggests how, for instance, paid users should be provided better recommendations when compared to free users. They design a set of metrics that can take this disparity into account and then measure fairness accordingly. Another interesting way to measure unfairness is the concept of envy-free fairness, which is achieved when no one user prefers another user's recommendations way more significantly than their own \cite{do2022online}. While these metrics can identify consumer-side bias to an extent, they come with some limitations.

Limitations of current metrics used to quantify consumer-side bias in recommendation systems include: (i) over-simplifying the meaning of fairness in RS, which employs various techniques like collaborative and content-based filtering and hybrid methods.
Simple notions can fail to capture disparities that exist across different types of items, for example, genres, when considering movie recommendations. Some metrics that can fall prey to this oversimplification issue include \cite{Weydemann19,NIPS2017_e6384711, Fu20,10.1145/3442381.3449866,islam21,foulds2019bayesianmodelingintersectionalfairness,9101635,do2022online,10.1145/3534678.3539269,Melchiorre21,deldjoo_flexible_2021};
(ii) not utilizing ranks when evaluating recommendation quality can yield considerable issues. 
This is due to the fact that recommendations are displayed one after another, so the items on higher ranks must be more relevant to keep the user satisfied. 
Thus, capturing the quality of recommendations using the ranks reflects a more complete way of evaluating models. 
Some metrics that can fall prey to this issue include \cite{NIPS2017_e6384711,Weydemann19,do2022online,10.1145/3534678.3539269,deldjoo_flexible_2021}.
It is important to state however, that even considering rank can give rise to positional bias, which refers to the tendency of users to favor the items that appear on top of a ranked list. 
Evaluating till a certain position like Recall@k can lead to a skewed sense of assessment of how the recommendation model performs. 
Additionally, metrics like MAP (Mean Average Precision), which normally treats relevance as a binary value (0: not relevant and 1: relevant) can also fail to capture the nuanced relevance that can come from items having multiple categories. 
So, using more than one metric (both with and without using ranks) to quantify bias is essential; 
(iii) relying only on one type of fairness metric can obscure underlying biases and give a false impression of fairness in recommendation systems. 

Hence, using multiple metrics can help uncover hidden biases. Additionally, a model that is fair according to one metric can fail to hold other fairness metrics and risk overlooking subtle unfairness issues.

We want to highlight some of the works that have taken into account different classes when evaluating recommendations \cite{lin2019crankvolumepreferencebias,10.1145/3240323.3240372,10.1145/3292500.3330691,10.1145/3626772.3657794,10068703,10.1145/3308560.3317595,10.1145/3038912.3052612,10.1145/3589334.3648158}.
For instance, \cite{lin2019crankvolumepreferencebias} groups users on certain attributes and items by category, then measures preference ratio, which is the fraction of liked items by a group across categories. Next, they measure the bias disparity by taking the preference and recommended ratios' relative differences. This is close to our work but still doesn't account for the ranks of items and we evaluate the direct comparison of the recommendations for males and females. Additionally, \cite{10.1145/3240323.3240372} introduce calibrated recommendations, ensuring the recommended items align with user preferences without overemphasizing particular categories. The work by \cite{10.1145/3292500.3330691} uses a measure \textit{Skew@k} to evaluate proportions of candidates based on sensitive attributes, and \cite{10.1145/3626772.3657794} uses a fairness metric called Attention Weighted Ranked Fairness (AWRF) \cite{10.1145/3308560.3317595} to ensure there is balance in exposure in different groups of providers. While both these works ensure group fairness, our work is more concentrated on evaluating the distribution of content categories for different groups. Unlike the work by \cite{10.1145/3626772.3657794} that focuses on provider-side fairness, we focus on consumer-side fairness. 
\section{Proposed Evaluation Metrics}

\subsection{Motivating Fairness Concern}
\label{subsec:motivating_example}

Our example is a typical offline setting recommendation system, which is trained using historical user and movie interactions. 
For our scenario, let us assume we have $u_1$, a male user who has watched numerous action and sci-fi movies, with some romance movies. 
We also have $u_2$, a female user who has watched a lot of drama movies but also a few action movies. 
Let us say we decide to calculate overall precision for each user group (male and female) and then compare them to ensure the model's fairness. 

\subsubsection{Potential Issue}
Machine learning systems tend to learn and amplify bias from the training data \cite{10.5555/3157382.3157584,lum2016,pmlr-v81-ensign18a,10.1145/3287560.3287572,kheya2024pursuitfairnessartificialintelligence}. Recommender systems are no different, as they can pick up on stereotypical user-item interactions and make biased recommendations based on sensitive attributes of users \cite{TsintzouPT19,Melchiorre21}. Biased recommendations can have a broad impact on users if the models recommend content just because it aligns with certain stereotypes, for instance, males like action movies, and females like romance movies. 

This imbalance in recommendation can lead to a situation in which users are only exposed to items that align with part of their preferences and stereotypical norms. This would potentially filter out diverse content and prevent users from discovering new movies. As time passes, $u_1$ might stop getting romance movies recommended to them, even though they enjoy them. 
For $u_2$, a similar case could arise for action movies. 
Essentially, this imbalance can trap the users in a bubble of recommendations with only their established preferences and gender-stereotypical genres. Additionally, if a user's established preferences are already aligned with gender stereotypes, then the bias in recommendation will intensify further giving rise to a filter bubble, with very redundant movie recommendations.

\subsubsection{Falling short when quantifying gender bias}
Moreover, as mentioned earlier, using some performance metrics for both genders and comparing them to evaluate fairness is not a great idea. 
The two groups can have similar scores, even if the model is making biased recommendations by choosing to neglect certain categories for certain users. 
Our proposed metrics address this issue by breaking down the recommendations by category to get a more nuanced sense of fairness.

The key takeaway here is the importance of learning fair user and item representations, as well as evaluating fairness in the output. 
Even if a model is trained only on user-item-rating interactions with no explicit mention of sensitive attributes, the model can still infer this private information due to the correlation between their behavior and their sensitive attributes \cite{bothmann2024fairnessroleprotectedattributes,10.1145/3106237.3106277,10.1145/3368089.3409697}. 
So, we have to take precautionary measures when training the model itself, so it is unable to learn these correlations. 
We also wanna discuss how it is important to ensure personalization, including gender-specific preferences to enhance user satisfaction, but such preferences should be balanced against any risk of reinforcing stereotypes. 

Please note in our study we focus on binary genders, acknowledging there are many other gender identities not represented here.

\subsection{Notation}

We present all metrics for fairness assessment using the following mathematical notation:
\begin{itemize}
    \item $u_i$: A single user, where $i$ indexes the users.
    \item $v_j$: A single item, where $j$ indexes the items.
    \item $\mathcal{U}$ and $\mathcal{V}$: The set of users and items, respectively.
    \item $\mathcal{U}_m$ and $\mathcal{U}_f$: The set of male and female users, respectively.
    \item $c$: An item category, such as Action, Sci-Fi, Romance, etc.
    \item $C$: A category matrix where $C_{j,c} = 1$ if category $c$ is associated with item $v_j$, and $C_{j,c} = 0$ otherwise.
    \item $C_{v_j}$: The list of categories associated with item $v_j$.
    \item $TopK_{u_i}$: The set of top $K$ recommended items for user $u_i$.
    \item $\mathrm{C}$: Represents the set of categories for items.
\end{itemize}

To assess fairness, we adapt and extend Information Retrieval metrics, introducing both non-ranking and ranking-based metrics, which are detailed in the following subsections.

\subsection{Non-ranking-based metrics}
The first set of metrics we propose evaluates the fairness of a recommender system without considering the ranking of movies. 
\subsubsection{Category Coverage (CC)} 
\label{subsec:gp}

This metric estimates the proportion of recommended items associated with category $c$ relative to all categories for all users $u_i$. 
This essentially captures the category-specific performance of the recommended items. 
It is defined as follows:

\begin{equation}
CC(c, \mathcal{U}) = \frac{1}{|\mathcal{U}|} \sum_{u_i \in \mathcal{U}} \frac{1}{|TopK_{u_i}|} \sum_{v_j \in TopK_{u_i}} \frac{C_{j,c}}{|C_{v_j}|}
\label{eq:M1}
\end{equation}

\subsubsection{Relative Category Representation (RCR)}
This metric estimates the proportion of category $c$ in recommended items relative to the proportions of all categories available from the whole dataset. 
This helps provide insights on category-specific items and is defined as follows: 

\begin{equation}
RCR(c, \mathcal{U}) = \frac{1}{|\mathcal{U}|} \sum_{u_i \in \mathcal{U}} \frac{\sum_{v_j \in TopK_{u_i}} \frac{C_{j,c}}{|C_{v_j}|}}{\sum_{v_k \in \mathcal{V}} \frac{C_{k,c}}{|C_{v_k}|}}
\label{eq:M2}
\end{equation}

These two metrics help us quantify how relevant a category is for items recommended to a given set of users. For instance, $CC$ would help us quantify the diversity of items recommended by reflecting the overall distribution of recommended content when computed for different $c$. Whereas, $RCR$ (calculated for different $c$) would measure if the recommendations suppress or amplify certain categories by comparing them to the actual distribution of available content.
Both $CC$ and $RCR$ provide a more nuanced understanding of how the recommendation system performs but don't take into account the position of the recommended items. 
For recommendation systems, where in most cases items surface one after another, it is vital that the items on the top of the menu are the most relevant. 
Hence, to ensure that ranking order is also considered for evaluating recommended items, in the next section we introduce rank-based metrics.

\subsection{Rank-based Metrics}
\label{sec:rbm}
We now introduce our metrics that utilize ranking to provide a comprehensive assessment of fairness.

\subsubsection{Category Mean Average Precision (CMAP)}
For a given set of users, this metric estimates the proportion of recommended items associated with category $c$ by incorporating the rank of the items. 
A category-specific average precision score is computed for each user, followed by averaging these scores across all users.
This metric is defined as follows:


\begin{equation}
CMAP(c, \mathcal{U}) = \frac{1}{|\mathcal{U}|} \sum_{u_i \in \mathcal{U}}
\frac{\sum_{j=1}^{|TopK_{u_i}|}  P(j) \cdot \frac{C_{j,c}}{|C_{v_j}|}}{\sum_{v_j \in \mathcal{V}} \frac{C_{j,c}}{|C_{v_j}|}}
\label{eq:M3}
\end{equation}


\noindent where $
P(j) = \frac{1}{j} \sum_{k=1}^{j} \frac{C_{k,c}}{|C_{v_k}|}
$.

\subsubsection{Category Discounted Cumulative Gain (CDCG)}
The score for this metric is discounted based on the position of the item in the recommended list. 
While similar to the previous metric, CDCG uses a logarithmic discount factor, which is higher for items that appear lower down in the list. It is defined as follows:

\begin{equation}
CDCG(c, \mathcal{U}) = \frac{1}{|\mathcal{U}|} \sum_{u_i \in \mathcal{U}} \frac{1}{|TopK_{u_i}|} \sum_{j=1}^{|TopK_{u_i}|} \frac{\frac{C_{j,c}}{|C_{v_j}|}}{\log(j + 1)}
\label{eq:M4}
\end{equation}

\subsubsection{Category Mean Reciprocal Rank (CMRR)}
This metric is based on MRR, which is essentially the mean reciprocal of the rank of the first relevant item. 
We adapt it for assessing fairness and we define it as follows:
\begin{equation}
CMRR(c, \mathcal{U}) = \frac{1}{|\mathcal{U}|} \sum_{u_i \in \mathcal{U}} \frac{1}{|TopK_{u_i}|} \sum_{j=1}^{|TopK_{u_i}|} \frac{\frac{C_{j,c}}{|C_{v_j}|}}{j}
\label{eq:M5}
\end{equation}

\subsubsection{Category RPrecision (CRP)}
This metric uses the category-specific precision but only for the top \(\lfloor R_c \rfloor\) results, where $R_c$ represents the proportion of category $c$ when considering all items in $\mathcal{V}$ (i.e., $R_c=\sum_{j=1}^{|\mathcal{V}|} \frac{C_{j,c}}{|C_{v_j}|}$, note that item $v_j$ can belong to multiple categories like movie genres). 
This metric is defined as follows:
\begin{equation}
CRP(c, \mathcal{U}) = \frac{1}{|\mathcal{U}|} \sum_{u_i \in \mathcal{U}} \frac{\sum_{j=1}^{|Top\lfloor R_c \rfloor_{u_i}|} \frac{C_{j,c}}{|C_{v_j}|}}{\lfloor R_c \rfloor}
\label{eq:M6}
\end{equation}

All four of the ranking-based metrics, take into consideration the proportions of categories and use rank as a discounting factor. The discounting factor is different for each of them and has distinct nuances. For instance, $CRP$ would only evaluate recommendations till a certain rank \(\lfloor R_c \rfloor\) to measure if the model is providing a proportional amount of recommendations for $c$ relative to its presence in the dataset. Another metric like $CDCG$ for example would capture something different, which is the relevance of $c$ by giving more weight to items (falling under category $c$) appearing higher in the list.

\subsection{Group Fairness (Gender)}
Gender Balance Score (GBS) for a given category $c$, is the absolute difference of category distribution values for $\mathcal{U}_m$ and $\mathcal{U}_f$ based on any of the category-aware metrics discussed above. 
This metric is an adaptation of \textit{demographic parity} as mentioned in Section \ref{subsec:fair_notion}. 
We want to make sure that the category distributions of the recommended items for the groups, of males and females are similar. 
For each metric $\mathcal{M}$, defined above, we want to ensure that the difference of values calculated for the groups male and female summed up for all categories are close to 0. 
This can be formalized as:
\begin{equation} \bigtriangleup
\mathcal{M}_c=| \mathcal{M}(c,\mathcal{U}_m) - \mathcal{M}(c,\mathcal{U}_f)|
\label{eq:delM}
 \end{equation}
\begin{equation} 
GBS(\mathcal{M}) =\sum_{c \in \mathrm{C}} \bigtriangleup
\mathcal{M}_c \approx 0
\label{eq:GBS}
 \end{equation}

\section{Genre Aware Regularization For Gender Fairness}
\label{sec:fairloss}

Now that we have introduced our set of metrics for capturing a nuanced sense of fairness for recommended items, we aim to learn fair representations that are category-aware for users of different genders.
To promote fairness in recommendation models we employ an in-processing regularization technique inspired by the approach first proposed by \cite{pmlr-v54-zafar17a}. 
We propose to incorporate Category Coverage (Section \ref{subsec:gp}) into the loss function alongside the primary recommendation loss, encouraging the model to optimize for category-aware fairness. We want to emphasize that while we select $CC$ as an example here, optimizing on any of the other metrics (as long as the implementation is differentiable) is possible.
We incorporate this fairness regularizer into the loss function of each baseline model—MF, VAE-CF, and NeuMF.

Specifically, to discourage the model from learning any biased representations, we incorporate the following regularizer:
\begin{equation}
\mathcal{L}_{FairGenreGender} = GBS(CC) = \sum_{c \in \mathrm{C}}|CC(c,\mathcal{U}_m)-CC(c,\mathcal{U}_f)|
 \label{eq:myloss}
\end{equation}
The goal is to minimize the difference in category distribution of the items being recommended to the users of each sensitive group. 
The new loss function for MF, VAE-CF and NeuCF models can be written as:
\begin{equation}
\small \mathcal{L} = \alpha \mathcal{L}_{FairGenreGender} + (1-\alpha) \mathcal{L}_{Recommendation}
\label{eq:fairequation}
\end{equation}

\noindent where \(\alpha\) is used to calibrate the trade-off between the model loss and the fairness term. \(\mathcal{L}_{Recommendation}\) represents the respective recommendation loss term of each model. By tuning this hyperparameter, we ensure that fairness does not overly compromise the recommender's ability to respect users' personal preferences (even if there are natural differences across liked categories for users of different genders). 
In order to modulate the gender loss value and to ensure it is able to make a significant contribution to the loss function, we pass it through a sigmoid function, centering at 0.5 and scaling it by 0.1. Additionally, the maximum batch loss value of the actual recommendation loss is used to scale the gender loss up to ensure both losses are on the same scale.
In the next section we show how using this fairness regularization helps the models minimize bias in recommendations provided to users of different genders for different categories.

\section{Experiments}
In the sub-sections below, we outline our experimental setup.
\subsection{Datasets}
\begin{table}[t]
\caption{Description of datasets.}
\label{tbl:dataset}
\resizebox{\columnwidth}{!}{%
\begin{tabular}{ |c|c|c|c|c|c| } 
\hline
Dataset &  Users & Item & Interactions & Categories for Items\\
 \hline
ML 100K \cite{10.1145/2827872} & 943 &  1,349  & 99,287 & 18 \\
ML 1M \cite{10.1145/2827872} & 6,040   & 3,416  & 999,611 & 18 \\
Yelp \cite{mansoury2019biasdisparitycollaborativerecommendation} & 1,316   & 1,272  & 97,991 & 21\\
\hline
\end{tabular}
}
\end{table}
Our experiments are performed on three recommendation datasets summarized in Table~\ref{tbl:dataset}.
The data is split into training, validation, and test sets with a 70:10:20 ratio following user-based split scheme.
It's important to highlight that items can fall under multiple categories. For clarity and ease of understanding, we focus on four representative categories—\textit{Action}, \textit{Romance}, \textit{Sci-Fi}, and \textit{Drama} from the MovieLens datasets and \textit{Coffee, Tea \& Desserts}, \textit{Arts \& Entertainment}, \textit{Travel \& Transportation} and \textit{Asian} from the yelp dataset — when visualizing our metric values; however all categories are included in our experiments.


\subsection{Evaluating Performance}
 For measuring the performance of recommendations made, we use HitRatio@k calculated for each user and then averaged. Additionally, we use a ranking-based metric NDCG@k (Normalized Discounted Cumulative Gain) which gives higher relevance to items appearing higher in the ranked list. This too, is calculated for each user using their top \(k\) recommendations and then averaged. We calculate these values for \(k=50\).



\subsection{Base Recommendation Models}
For analyzing which models manifest gender bias we evaluate several recommendation approaches. When selecting these models, we include a variety of algorithms, ranging from traditional ones to more modern ones including Matrix Factorization, UserKNN, ItemKNN, NeuMF and VAE-CF (more detailed information on these and the fairness-aware models can be found in Section \ref{sec:baseline}).

Additionally, we include a regularization-based fairness-aware model BeyondParity \cite{NIPS2017_e6384711} and a counterfactually fair recommendation model SM-GBiasedMF \cite{li2021towards}. To weigh the fairness-aware loss for SM-GBiasedMF, we use \(\lambda=20\) for all experiments since this value seems to work well as mentioned in \cite{li2021towards}. To make a fair comparison, we use BeyondParity's \(U_{val}\) (refer to Equation \ref{eq:bp}) as a regularization term in the same setting as shown in Equation~\ref{eq:fairequation}, with the same value of alpha we use for our MF model. We use an early stopping strategy for the three models being compared, monitoring NDCG@20 with a delta of 0.0005 and patience of 10 epochs with a maximum number of iterations of 50 and 100 for the 100K and 1M datasets, respectively.

We utilize the Cornac framework \cite{JMLR:v21:19-805}, with customizations to meet our requirements, including the implementations of differentiable ways to obtain top k category distribution for calculating our gender loss term for different models. When calculating our regularization term, we generate a top \(k\) recommendation list of size \(k=50\). Additionally, all our results for all metrics for GBS (except CRP) use the top 50 recommended items. For GBS(CRP) we use top \(\lfloor R_g \rfloor\) as explained for Equation \ref{eq:M6}.

\section{Analysis of the proposed metrics}
In this section, we discuss the results we obtained from our experiments through a comprehensive analysis.
\subsection{Baselines Comparison}
\begin{table}[]
\centering
\resizebox{\columnwidth}{!}{%
\begin{tabular}{|ll|ll|llllll|}
\hline
Model & &\begin{tabular}[c]{@{}l@{}}Hit\\ Ratio\end{tabular}$\uparrow$ & NDCG$\uparrow$ & CC$\downarrow$ & RCR$\downarrow$ & CMAP$\downarrow$ & CDCG$\downarrow$ & CMRR$\downarrow$ & CRP$\downarrow$ \\ 
\hline

\multicolumn{10}{|c|}{{\bf ML 100k Dataset}} \\
\hline

\multicolumn{2}{|l|}{UserKNN}               &  0.4931& 0.0321 & 0.0283     & 0.0340  &  0.0029& 0.0063 & 0.0017 & \multicolumn{1}{l|}{0.0178}       \\ \hline
\multicolumn{2}{|l|}{ItemKNN}          &     0.4698             & 0.0347 & 0.0675  & 0.0749  & 0.0095 & 0.0181 & 0.0070 &  \multicolumn{1}{l|}{0.0567}    \\ \hline
\multirow{2}{*}{MF}          & Original                                            &   \textbf{0.9003}  &         \textbf{0.1979}   &  0.0288    &    0.0321   &   0.0043  &  0.0089    & \textbf{ 0.0055}   & \multicolumn{1}{l|}{0.0276}       \\ \cline{3-10} 
                             & Fair                                                        & 0.8950  & 0.1794    &\textbf{ 0.0228}      & \textbf{0.0312 }     & \textbf{0.0037 }     &\textbf{ 0.0085}      & 0.0060     & \multicolumn{1}{l|}{\textbf{0.0234}}         \\ \hline
\multirow{2}{*}{VAE-CF}      & Original   &   0.8749             & \textbf{0.1643} & 0.0187  & 0.0210  & 0.0030 & 0.0052 & \textbf{0.0026} & \multicolumn{1}{l|}{0.0137}     \\ \cline{3-10} 
                             & Fair          & \textbf{0.8759}             & 0.1575 & \textbf{0.0140 } & \textbf{0.0186}  & \textbf{0.0026}& \textbf{0.0047} & 0.0037 & \multicolumn{1}{l|}{\textbf{0.0128 } }      \\ \hline
\multirow{2}{*}{NEU-MF}      & Original      & \textbf{0.9470}  & \textbf{0.2402} & 0.1723  & 0.1452  & 0.0268 & 0.0502 & 0.0220 & \multicolumn{1}{l|}{0.1411 }      \\ \cline{3-10} 
                             & Fair          & 0.9194             & 0.2305 & \textbf{0.0333  }& \textbf{0.0405}  & \textbf{0.0056} & \textbf{0.0093} & \textbf{0.0059} & \multicolumn{1}{l|}{\textbf{0.0342}  }     \\ \hline
\multicolumn{10}{|c|}{{\bf ML 1M Dataset}} \\ 
\hline
\multicolumn{2}{|l|}{UserKNN}                &   0.3997   &  0.0179  &  0.0204   &  0.0110&       0.0008 &0.0043&0.0008& \multicolumn{1}{l|}{0.0174}   \\ \hline
\multicolumn{2}{|l|}{ItemKNN}      &    0.3565   &0.0383      & 0.0881    & 0.0351 & 0.0051  &  0.0224 & 0.0082 & \multicolumn{1}{l|}{0.0650}    \\ \hline
\multirow{2}{*}{MF}          & Original      &\textbf{ 0.8485}                                               &   \textbf{0.1522}    &   0.1775   &      0.0486 &  0.0097   &  0.0495    &  0.0204   & \multicolumn{1}{l|}{0.1440}       \\ \cline{3-10} 
                             & Fair          &    0.8055  & 0.1158   &  \textbf{0.1160}   &   \textbf{0.0314 }  &    \textbf{0.0067}  &   \textbf{0.0344}  & \textbf{0.0162}   & \multicolumn{1}{l|}{\textbf{0.0760}}         \\ \hline
                             \multirow{2}{*}{VAE-CF}      & Original    &  \textbf{0.8315 }  & \textbf{0.1473} & 0.2551  & 0.0708  & 0.0154 & 0.0706 & 0.0282 &\multicolumn{1}{l|}{0.1900  }    \\ \cline{3-10} 
                             & Fair          & 0.8280 &  0.1423 &  \textbf{0.2229 } & \textbf{0.0637}  & \textbf{0.0132} & \textbf{0.0631} &  \textbf{0.0263}&\multicolumn{1}{l|}{\textbf{0.1502}}       \\ \hline
\multirow{2}{*}{NEU-MF}      & Original      & \textbf{ 0.9108 }   &  0.1336   &   0.3096  & 0.0998 &  0.0246  & 0.0835  & 0.0316 & \multicolumn{1}{l|}{0.2180  }   \\ \cline{3-10} 
                             & Fair   &          0.8333 & \textbf{0.1359}                                                       & \textbf{0.1503}     & \textbf{0.0415}     & \textbf{0.0086}      &\textbf{ 0.0436 }     & \textbf{0.0191 }     & \multicolumn{1}{l|}{\textbf{0.0959}}       \\ \hline
\multicolumn{10}{|c|}{{\bf Yelp Dataset}} \\ 
\hline

\multicolumn{2}{|l|}{UserKNN}                &   0.4886    &  0.0312    &  0.0068   & 0.0252 &   0.0068    &0.0026&0.0022& \multicolumn{1}{l|}{0.0129}   \\ \hline
\multicolumn{2}{|l|}{ItemKNN}      &    0.5030    &    0.0344   &   0.0242  & 0.0513 &   0.0060 & 0.0065  & 0.0035 & \multicolumn{1}{l|}{0.0219}    \\ \hline
\multirow{2}{*}{MF}          & Original      &     \textbf{0.8587 }                                          &   \textbf{0.1209 }  &   0.0358   &0.0342&0.0055&0.0088&0.0036& \multicolumn{1}{l|}{0.0296}       \\ \cline{3-10} 
                             & Fair          & 0.7500 &  0.0773  &  \textbf{0.0159 }  &\textbf{0.0312}&    \textbf{0.0019 } & \textbf{0.0042}&\textbf{0.0019}& \multicolumn{1}{l|}{\textbf{0.0130}}         \\ \hline
                             \multirow{2}{*}{VAE-CF}      & Original      &  \textbf{0.8131 }                                                  & \textbf{0.0985 }   & 0.0045       & 0.0041      & 0.0003      & 0.0007      & 0.0003      & \multicolumn{1}{l|}{0.0018}       \\ \cline{3-10} 
                             & Fair          &  0.8032                                                 & 0.0961   &   \textbf{0.0021}   & \textbf{0.0019}      & \textbf{0.0002 }      &  \textbf{0.0004}    &   \textbf{0.0001}     & \multicolumn{1}{l|}{ \textbf{0.0011}}       \\ \hline
\multirow{2}{*}{NEU-MF}      & Original      &  \textbf{0.9347 }                                                & \textbf{0.1794}     & 0.0592       & 0.0850      & 0.0134    & 0.0152      & 0.0057    & \multicolumn{1}{l|}{0.0651}       \\ \cline{3-10} 
                             & Fair          &  0.8533  &  0.1124    & \textbf{0.0309 }      & \textbf{ 0.0401 }    & \textbf{0.0054 }    & \textbf{0.0079}      &  \textbf{0.0028}      & \multicolumn{1}{l|}{ \textbf{0.0394}}       \\ \hline
\end{tabular}
}
\caption{Performance and Bias Evaluation values (GBSs) for 5 baseline models, along with the fair models. }
    \label{tab:baselines}
\end{table}

\begin{figure*}[h]
    \centering
    \includegraphics[width=1\textwidth]{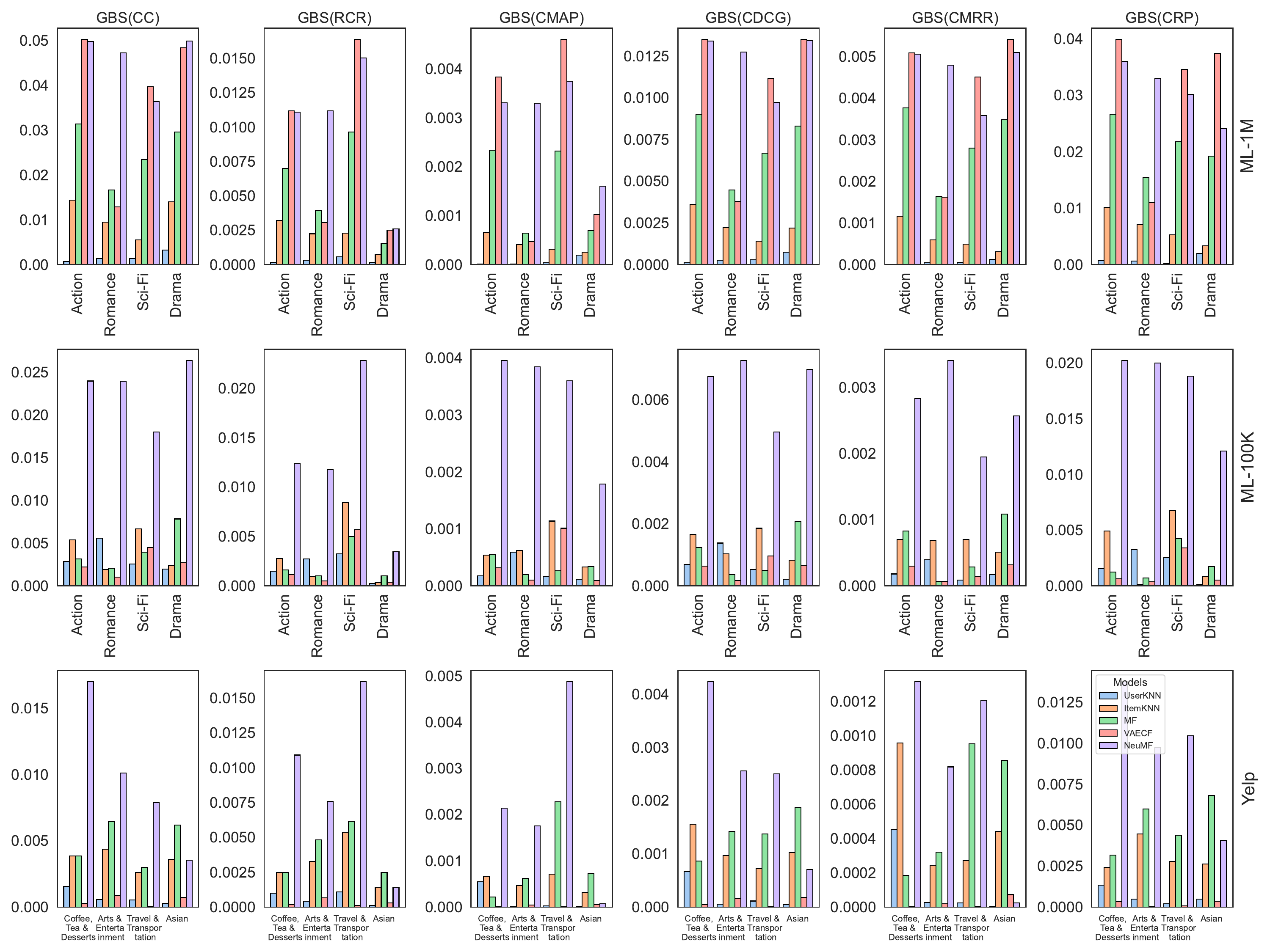}
    \caption{ Comparison of Bias values for six of our metrics for all three dataset.}
    \label{fig:baseline_100k}
\end{figure*}

\subsubsection{Bias Evaluation}
\begin{figure}[h]
    \centering
    \includegraphics[width=0.5\textwidth]{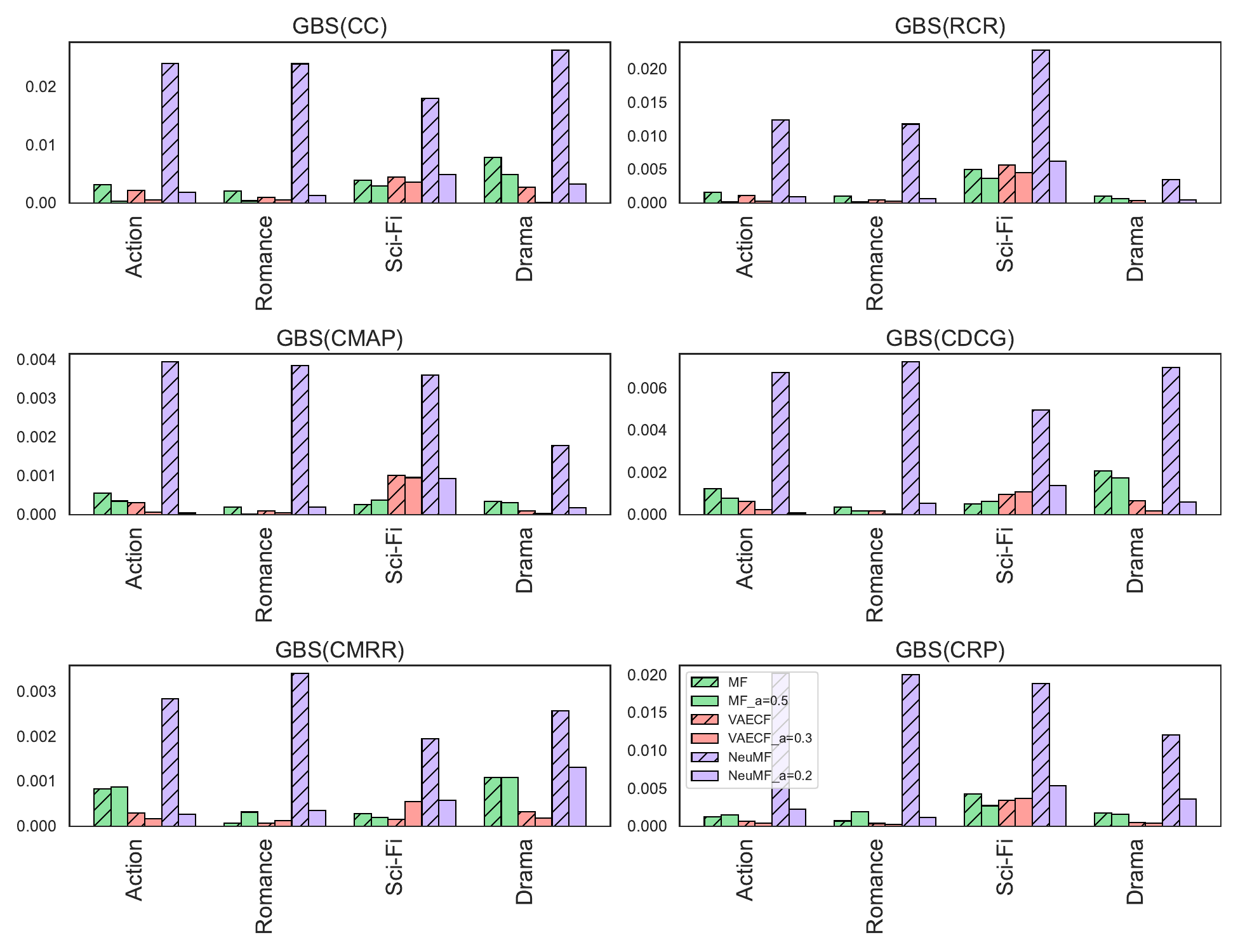}
    \caption{Reduction in bias scores after using our fairness-aware regularizer. Since all datasets provide similar outcomes, we present the results for only the ML 100K dataset. }
    \label{fig:models_reg_100k}
\end{figure}

We report our results in Figures \ref{fig:baseline_100k}, \ref{fig:models_reg_100k}, and Table \ref{tab:baselines}. 
For the KNN-based models, ItemKNN seems more bias-prone. This can be due to the nature of the model of calculating similarity between items based on the interactions by the users. For
instance, in the ML-100k dataset, the average rating for romance
movies is higher for female users than for males. This kind of imbalance can reinforce category-related gender bias when items are recommended. 
 Among the three other baselines, the MF-based models can be considered more biased. For the MF model, explicit embeddings are used to store user and item representations, which are likely to capture correlations of user behavior (e.g., liking certain genres) and their sensitive attributes. For VAE-CF, variational autoencoders are used to learn probabilistic latent representations of users, which are still sensitive to capturing biased representations but not as much as NeuMF. NeuMF captures both linear and non-linear relationships between users and items through Generalized Matrix Factorization (GMF) and Multi-Layer Perceptrons (MLP). So, NeuMF can capture biases in the data, especially the intricate patterns about user preferences, which can reflect social stereotypes. The MF model itself is not as complex as NeuMF, so our results of NeuMF being more biased than MF is reasonable.
 For the two smaller datasets, VAE-CF doesn’t manifest high values for bias, which can be due to the model not being able to fully capture representations of gendered preferences because of the lower number of interactions available. Comparatively, for the 1M dataset, the model can learn the stereotypical patterns available in the dataset more effectively and reinforce these biases in the latent representations. In general, all models manifest more gender bias for the larger dataset than the smaller ones.
 For all baseline models, the Gender Balance Scores are \(\not \approx 0\), which highlights the fact that they are producing recommendations in a discriminative way where they choose certain genres to be relevant for certain groups of users. From Table \ref{tab:baselines}, GBSs are higher for \(CC\) or \(RCR\) (ref to Equations: \ref{eq:M1} and \ref{eq:M2}). These two are classification-based metrics and do not consider ranks, so there is no discounting done in the value of any item that appears lower in the list but is still relevant (category-wise). GBS(CMRR) has the lowest values for almost all models since it has a strong discounting factor for items that appear lower in rank. 

The bias reduction after using the fairness aware loss term is displayed in Figure \ref{fig:models_reg_100k}. We can see a notable improvement in the bias scores, especially for the NeuMF model. The regularizer term is most effective for reducing bias in the NeuMF model, with an average decrease across all six metrics of 77\%, 53\%, and 50\% for ML100K, ML1M, and Yelp datasets, respectively. The impact of the fairness regularizer is less pronounced for the VAE-CF model. We believe this is due to the model's nature of learning probabilistic latent factors for users and items that don't amplify gender bias like the MF models which use embeddings to learn user and item interactions. As a result, the regularizer has less impact on the model's learning process.

\subsubsection{Performance Evaluation}\label{sec:perform-evaluation}
The performance drop for VAE-CF and NeuMF is less than that of the MF model, however, it all depends on the selection of $\alpha$ (more on this in Section \ref{sec:abalation}). For all models, when choosing $\alpha$ we ensure there is substantial bias reduction without too significant of a drop in performance. 
In some cases the fair models outperform the original models, this although seems counter-intuitive, but is presumably due to the fairness term acting as a regularization term which can help reduce over-fitting in the model to some extent (as noted in \cite{keya2020equitableallocationhealthcareresources,10.1145/3442381.3449904}).

\subsection{Fairness-Aware Baselines}
\begin{table}[h]
\centering

\resizebox{\columnwidth}{!}{%
\begin{tabular}{|c|cc|cccccc|}
\hline
Model & NDCG$\uparrow$ & HitRatio$\uparrow$ & CC$\downarrow$  & RCR$\downarrow$  & CMAP$\downarrow$  & CDCG$\downarrow$  & CMRR$\downarrow$  & CRP$\downarrow$  \\
\hline
\multicolumn{9}{|c|}{{\bf ML 100K Dataset}} \\
\hline
MF$_{a=0.5}$ & \textbf{0.1794} &    \textbf{0.8950} & \textbf{0.0228} &  \textbf{0.0312} & \textbf{ 0.0037} & \textbf{ 0.0085} &  0.0060 & \textbf{ 0.0234 }\\

SM-GBiasedMF   & 0.1230 & 0.8537 & 0.03785 & 0.0378& 0.0059  & 0.0114 & \textbf{0.0053} &   0.0286    \\
BeyondParity     & 0.0790&  0.6384& 0.0528&0.0752&0.0093 & 0.0147 &  0.0068&  0.0480 \\
\hline
\multicolumn{9}{|c|}{{\bf ML 1M Dataset}} \\
\hline
MF$_{a=0.6}$ & \textbf{0.1158 }  & 0.8055  &\textbf{0.1160} &\textbf{0.0314 }&  \textbf{0.0067}&\textbf{0.0344} &0.0162 &\textbf{0.0760}\\
SM-GBiasedMF    &0.0846   &  \textbf{0.8075} &0.2590 &0.0726 & 0.0152 & 0.0656& 0.0211 & 0.2058\\
BeyondParity   &  0.0829   &     0.6495   & 0.1307&0.0539 & 0.0079 & 0.0369&\textbf{0.0154}&0.0838\\

\hline
\multicolumn{9}{|c|}{{\bf Yelp Dataset}} \\
\hline
MF$_{a=0.3}$ & 0.0773&0.7500  &  0.0159   &\textbf{0.0312}&    \textbf{0.0019}  & 0.0042&\textbf{0.0019} &\textbf{0.0130} \\

SM-GBiasedMF    & \textbf{0.9157}  & \textbf{ 0.1343} &0.0515 &0.0789&0.0132&0.0140 & 0.0055 & 0.0587 \\
BeyondParity   & 0.0358    & 0.4901  & \textbf{0.0112}& 0.0342& 0.0066 & \textbf{0.0033}&0.0027 & 0.0241\\

\hline
\end{tabular}
}
\caption{Performance and Bias Evaluation Metric values for fairness-aware models.}
\label{tab:fairbaselines}

\end{table}

\begin{figure}[h]
    \centering
    \includegraphics[width=0.5\textwidth]{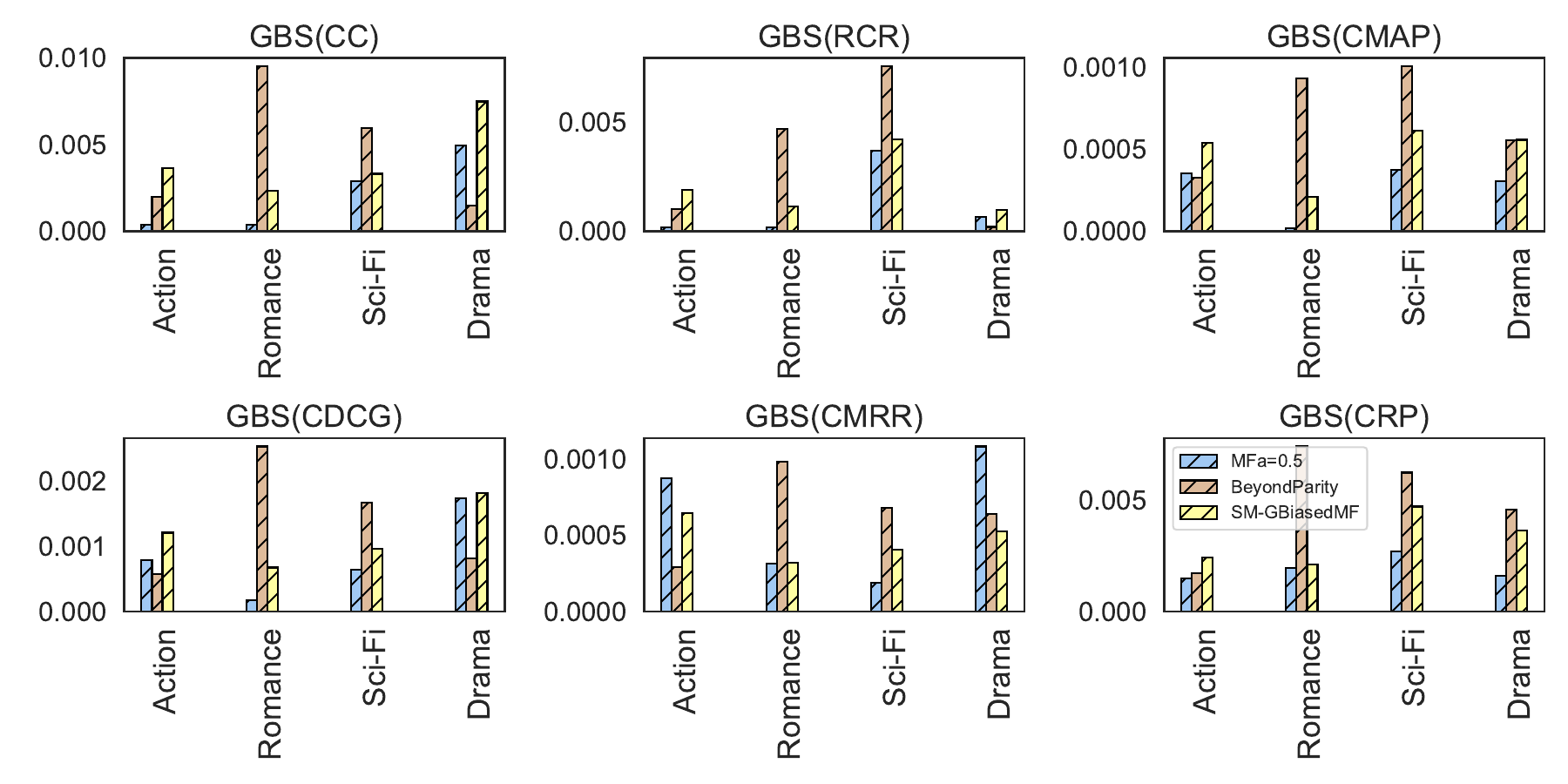}
    \caption{ Bias score for the fairness-aware models over four stereotypical genres for the ML 100K dataset.}
    \label{fig:compare_genre_fair}
\end{figure}

We can observe the results for our MF fair model when compared with two other fairness-aware models in Table \ref{tab:fairbaselines} and Figure \ref{fig:compare_genre_fair}. We chose to use our MF fair model since the other two fairness-aware models use MF as their foundational models. Our model has outperformed the other models in terms of bias scores across the majority of the comparisons. While our models don't excel in terms of performance, they still offer a better balance between performance and bias scores when compared with the other fairness-aware baselines. It is important to highlight that BeyondParity uses plain MSE loss, while the other two use BPR loss with negative sampling, which explains the exceptionally low-performance values. While we emphasize the significance of comparing the aggregated differences for each metric $\mathcal{M}$ for all the categories, we also want to highlight the importance of evaluating bias values separately. For instance, although GBS(CMRR) of SM-GBiasedMF is lower for the ML100K dataset when compared to ours, there is still significant bias for movies recommended that belong to the genres like \textit{Romance} and \textit{Sci-Fi} as observed in Figure \ref{fig:compare_genre_fair}. This strengthens our idea of having a set of metrics to quantify bias in a nuanced way since the model seems "fair" when considering overall scores for all categories, but closer inspection reveals how it is biased against certain categories. It is worth mentioning that SM-GBiasedMF is more biased than the plain MF model for the Yelp dataset. Our theory is that since this model needs more epochs to satisfy the early stopping criterion (as mentioned before), it likely over-fits the dataset, in turn amplifying the biases present in it.




\section{Discussion and Conclusion}
In this paper, we identify the underlying issues of current metrics for evaluating consumer-side bias in recommendation systems. To better quantify bias, specifically gender bias, in such models, we propose a set of metrics. These metrics help capture a nuanced sense of fairness on recommended items by considering categories of the items. Next, we demonstrate how introducing one of our metrics as a fairness loss term along with the recommendation loss helped minimize the unfairness manifested in models in terms of different categories of items recommended with minimal performance loss. Experiments on three real-world datasets using a variety of recommendation models, including fairness-aware models, show the effectiveness of our metrics in capturing bias. Additionally, after incorporating our loss term, the bias in the models was significantly reduced, with a favorable balance between fairness and accuracy. Our loss function is very flexible which allows for context-sensitive fairness. In domains such as job recommendation and housing, where fairness is vital, the functions can prioritize fairness. Conversely, for areas like movie recommendations it can balance personalization and fairness (ensuring user preferences are respected).
Our work aims to spread awareness about how simple metrics that are currently utilized to evaluate bias might be giving researchers a false sense of fairness, and a more refined approach like ours is required to address this issue. 
Our metrics are very versatile and can be easily adapted to measure provider-side fairness by utilizing proportions of item brands, for example, and can help quantify popularity bias in recommendations. Extending on this, the metrics can be used for measuring CP-fairness since we can measure bias from both the consumer side and provider side, by taking differences in values for different item providers for different demographic groups. Plus we plan to extend our metrics to consider sensitive attributes which are multi-valued, by using a pairwise difference scheme. 



\begin{acks}
This material is based upon work supported by the Air Force Office of Scientific Research under award number FA2386-23-1-4003.
\end{acks}

\bibliographystyle{ACM-Reference-Format}
\bibliography{mybib}


\begin{thebibliography}{81}


\ifx \showCODEN    \undefined \def \showCODEN     #1{\unskip}     \fi
\ifx \showISBNx    \undefined \def \showISBNx     #1{\unskip}     \fi
\ifx \showISBNxiii \undefined \def \showISBNxiii  #1{\unskip}     \fi
\ifx \showISSN     \undefined \def \showISSN      #1{\unskip}     \fi
\ifx \showLCCN     \undefined \def \showLCCN      #1{\unskip}     \fi
\ifx \shownote     \undefined \def \shownote      #1{#1}          \fi
\ifx \showarticletitle \undefined \def \showarticletitle #1{#1}   \fi
\ifx \showURL      \undefined \def \showURL       {\relax}        \fi
\providecommand\bibfield[2]{#2}
\providecommand\bibinfo[2]{#2}
\providecommand\natexlab[1]{#1}
\providecommand\showeprint[2][]{arXiv:#2}

\bibitem[Abdollahpouri et~al\mbox{.}(2019)]%
        {abdollahpouri2019unfairness}
\bibfield{author}{\bibinfo{person}{Himan Abdollahpouri}, \bibinfo{person}{Masoud Mansoury}, \bibinfo{person}{Robin Burke}, {and} \bibinfo{person}{Bamshad Mobasher}.} \bibinfo{year}{2019}\natexlab{}.
\newblock \showarticletitle{The unfairness of popularity bias in recommendation}.
\newblock \bibinfo{journal}{\emph{arXiv preprint arXiv:1907.13286}} (\bibinfo{year}{2019}).
\newblock


\bibitem[Abdollahpouri et~al\mbox{.}(2021)]%
        {10.1145/3450613.3456821}
\bibfield{author}{\bibinfo{person}{Himan Abdollahpouri}, \bibinfo{person}{Masoud Mansoury}, \bibinfo{person}{Robin Burke}, \bibinfo{person}{Bamshad Mobasher}, {and} \bibinfo{person}{Edward Malthouse}.} \bibinfo{year}{2021}\natexlab{}.
\newblock \showarticletitle{User-centered Evaluation of Popularity Bias in Recommender Systems}. In \bibinfo{booktitle}{\emph{Proceedings of the 29th ACM Conference on User Modeling, Adaptation and Personalization}} (Utrecht, Netherlands) \emph{(\bibinfo{series}{UMAP '21})}. \bibinfo{publisher}{Association for Computing Machinery}, \bibinfo{address}{New York, NY, USA}, \bibinfo{pages}{119–129}.
\newblock
\showISBNx{9781450383660}
\href{https://doi.org/10.1145/3450613.3456821}{doi:\nolinkurl{10.1145/3450613.3456821}}


\bibitem[Acharyya et~al\mbox{.}(2020)]%
        {Acharyya_Das_Chattoraj_Tanveer_2020}
\bibfield{author}{\bibinfo{person}{Rupam Acharyya}, \bibinfo{person}{Shouman Das}, \bibinfo{person}{Ankani Chattoraj}, {and} \bibinfo{person}{Md.~Iftekhar Tanveer}.} \bibinfo{year}{2020}\natexlab{}.
\newblock \showarticletitle{FairyTED: A Fair Rating Predictor for TED Talk Data}.
\newblock \bibinfo{journal}{\emph{Proceedings of the AAAI Conference on Artificial Intelligence}} \bibinfo{volume}{34}, \bibinfo{number}{01} (\bibinfo{date}{Apr.} \bibinfo{year}{2020}), \bibinfo{pages}{338--345}.
\newblock
\href{https://doi.org/10.1609/aaai.v34i01.5368}{doi:\nolinkurl{10.1609/aaai.v34i01.5368}}


\bibitem[Bolukbasi et~al\mbox{.}(2016)]%
        {10.5555/3157382.3157584}
\bibfield{author}{\bibinfo{person}{Tolga Bolukbasi}, \bibinfo{person}{Kai-Wei Chang}, \bibinfo{person}{James Zou}, \bibinfo{person}{Venkatesh Saligrama}, {and} \bibinfo{person}{Adam Kalai}.} \bibinfo{year}{2016}\natexlab{}.
\newblock \showarticletitle{Man is to computer programmer as woman is to homemaker? debiasing word embeddings}. In \bibinfo{booktitle}{\emph{Proceedings of the 30th International Conference on Neural Information Processing Systems}} (Barcelona, Spain) \emph{(\bibinfo{series}{NIPS'16})}. \bibinfo{publisher}{Curran Associates Inc.}, \bibinfo{address}{Red Hook, NY, USA}, \bibinfo{pages}{4356–4364}.
\newblock
\showISBNx{9781510838819}


\bibitem[Boratto et~al\mbox{.}(2022)]%
        {boratto2022consumer}
\bibfield{author}{\bibinfo{person}{Ludovico Boratto}, \bibinfo{person}{Gianni Fenu}, \bibinfo{person}{Mirko Marras}, {and} \bibinfo{person}{Giacomo Medda}.} \bibinfo{year}{2022}\natexlab{}.
\newblock \showarticletitle{Consumer fairness in recommender systems: Contextualizing definitions and mitigations}. In \bibinfo{booktitle}{\emph{European Conference on Information Retrieval}}. Springer, \bibinfo{pages}{552--566}.
\newblock


\bibitem[Bothmann et~al\mbox{.}(2024)]%
        {bothmann2024fairnessroleprotectedattributes}
\bibfield{author}{\bibinfo{person}{Ludwig Bothmann}, \bibinfo{person}{Kristina Peters}, {and} \bibinfo{person}{Bernd Bischl}.} \bibinfo{year}{2024}\natexlab{}.
\newblock \bibinfo{title}{What Is Fairness? On the Role of Protected Attributes and Fictitious Worlds}.
\newblock
\showeprint[arxiv]{2205.09622}~[cs.LG]
\urldef\tempurl%
\url{https://arxiv.org/abs/2205.09622}
\showURL{%
\tempurl}


\bibitem[Breese et~al\mbox{.}(1998)]%
        {10.5555/2074094.2074100}
\bibfield{author}{\bibinfo{person}{John~S. Breese}, \bibinfo{person}{David Heckerman}, {and} \bibinfo{person}{Carl Kadie}.} \bibinfo{year}{1998}\natexlab{}.
\newblock \showarticletitle{Empirical analysis of predictive algorithms for collaborative filtering}. In \bibinfo{booktitle}{\emph{Proceedings of the Fourteenth Conference on Uncertainty in Artificial Intelligence}} (Madison, Wisconsin) \emph{(\bibinfo{series}{UAI'98})}. \bibinfo{publisher}{Morgan Kaufmann Publishers Inc.}, \bibinfo{address}{San Francisco, CA, USA}, \bibinfo{pages}{43–52}.
\newblock
\showISBNx{155860555X}


\bibitem[Burke(2017)]%
        {burke_multisided_2017}
\bibfield{author}{\bibinfo{person}{Robin Burke}.} \bibinfo{year}{2017}\natexlab{}.
\newblock \bibinfo{title}{Multisided {Fairness} for {Recommendation}}.
\newblock
\href{https://doi.org/10.48550/arXiv.1707.00093}{doi:\nolinkurl{10.48550/arXiv.1707.00093}}
\newblock
\shownote{arXiv:1707.00093 [cs]}.


\bibitem[Chakraborty et~al\mbox{.}(2020)]%
        {10.1145/3368089.3409697}
\bibfield{author}{\bibinfo{person}{Joymallya Chakraborty}, \bibinfo{person}{Suvodeep Majumder}, \bibinfo{person}{Zhe Yu}, {and} \bibinfo{person}{Tim Menzies}.} \bibinfo{year}{2020}\natexlab{}.
\newblock \showarticletitle{Fairway: a way to build fair ML software}. In \bibinfo{booktitle}{\emph{Proceedings of the 28th ACM Joint Meeting on European Software Engineering Conference and Symposium on the Foundations of Software Engineering}} (Virtual Event, USA) \emph{(\bibinfo{series}{ESEC/FSE 2020})}. \bibinfo{publisher}{Association for Computing Machinery}, \bibinfo{address}{New York, NY, USA}, \bibinfo{pages}{654–665}.
\newblock
\showISBNx{9781450370431}
\href{https://doi.org/10.1145/3368089.3409697}{doi:\nolinkurl{10.1145/3368089.3409697}}


\bibitem[Chouldechova(2016)]%
        {Chouldechova2016FairPW}
\bibfield{author}{\bibinfo{person}{Alexandra Chouldechova}.} \bibinfo{year}{2016}\natexlab{}.
\newblock \showarticletitle{Fair prediction with disparate impact: A study of bias in recidivism prediction instruments}.
\newblock \bibinfo{journal}{\emph{Big data}}  \bibinfo{volume}{5 2} (\bibinfo{year}{2016}), \bibinfo{pages}{153--163}.
\newblock
\urldef\tempurl%
\url{https://api.semanticscholar.org/CorpusID:1443041}
\showURL{%
\tempurl}


\bibitem[Datta et~al\mbox{.}(2015)]%
        {datta15}
\bibfield{author}{\bibinfo{person}{Amit Datta}, \bibinfo{person}{Michael Tschantz}, {and} \bibinfo{person}{Anupam Datta}.} \bibinfo{year}{2015}\natexlab{}.
\newblock \showarticletitle{Automated Experiments on Ad Privacy Settings}.
\newblock \bibinfo{journal}{\emph{Proceedings on Privacy Enhancing Technologies}}  \bibinfo{volume}{1} (\bibinfo{date}{04} \bibinfo{year}{2015}).
\newblock
\href{https://doi.org/10.1515/popets-2015-0007}{doi:\nolinkurl{10.1515/popets-2015-0007}}


\bibitem[De-Arteaga et~al\mbox{.}(2019)]%
        {10.1145/3287560.3287572}
\bibfield{author}{\bibinfo{person}{Maria De-Arteaga}, \bibinfo{person}{Alexey Romanov}, \bibinfo{person}{Hanna Wallach}, \bibinfo{person}{Jennifer Chayes}, \bibinfo{person}{Christian Borgs}, \bibinfo{person}{Alexandra Chouldechova}, \bibinfo{person}{Sahin Geyik}, \bibinfo{person}{Krishnaram Kenthapadi}, {and} \bibinfo{person}{Adam~Tauman Kalai}.} \bibinfo{year}{2019}\natexlab{}.
\newblock \showarticletitle{Bias in Bios: A Case Study of Semantic Representation Bias in a High-Stakes Setting}. In \bibinfo{booktitle}{\emph{Proceedings of the Conference on Fairness, Accountability, and Transparency}} \emph{(\bibinfo{series}{FAT* '19})}. \bibinfo{publisher}{Association for Computing Machinery}, \bibinfo{pages}{120–128}.
\newblock
\showISBNx{9781450361255}
\href{https://doi.org/10.1145/3287560.3287572}{doi:\nolinkurl{10.1145/3287560.3287572}}


\bibitem[Deldjoo et~al\mbox{.}(2021)]%
        {deldjoo_flexible_2021}
\bibfield{author}{\bibinfo{person}{Yashar Deldjoo}, \bibinfo{person}{Vito~Walter Anelli}, \bibinfo{person}{Hamed Zamani}, \bibinfo{person}{Alejandro Bellogín}, {and} \bibinfo{person}{Tommaso Di~Noia}.} \bibinfo{year}{2021}\natexlab{}.
\newblock \showarticletitle{A flexible framework for evaluating user and item fairness in recommender systems}.
\newblock \bibinfo{journal}{\emph{User Modeling and User-Adapted Interaction}} \bibinfo{volume}{31}, \bibinfo{number}{3} (\bibinfo{date}{July} \bibinfo{year}{2021}), \bibinfo{pages}{457--511}.
\newblock
\showISSN{1573-1391}
\href{https://doi.org/10.1007/s11257-020-09285-1}{doi:\nolinkurl{10.1007/s11257-020-09285-1}}


\bibitem[Do et~al\mbox{.}(2022)]%
        {do2022online}
\bibfield{author}{\bibinfo{person}{Virginie Do}, \bibinfo{person}{Sam Corbett-Davies}, \bibinfo{person}{Jamal Atif}, {and} \bibinfo{person}{Nicolas Usunier}.} \bibinfo{year}{2022}\natexlab{}.
\newblock \showarticletitle{Online certification of preference-based fairness for personalized recommender systems}. In \bibinfo{booktitle}{\emph{Proceedings of the AAAI Conference on Artificial Intelligence}}, Vol.~\bibinfo{volume}{36}. \bibinfo{pages}{6532--6540}.
\newblock


\bibitem[Edizel et~al\mbox{.}(2020)]%
        {Edizel20}
\bibfield{author}{\bibinfo{person}{Bora Edizel}, \bibinfo{person}{Francesco Bonchi}, \bibinfo{person}{Sara Hajian}, \bibinfo{person}{Andre Panisson}, {and} \bibinfo{person}{Tamir Tassa}.} \bibinfo{year}{2020}\natexlab{}.
\newblock \showarticletitle{FaiRecSys: mitigating algorithmic bias in recommender systems}.
\newblock \bibinfo{journal}{\emph{International Journal of Data Science and Analytics}}  \bibinfo{volume}{9} (\bibinfo{date}{03} \bibinfo{year}{2020}), \bibinfo{pages}{197--213}.
\newblock
\href{https://doi.org/10.1007/s41060-019-00181-5}{doi:\nolinkurl{10.1007/s41060-019-00181-5}}


\bibitem[El~Halabi et~al\mbox{.}(2020)]%
        {NEURIPS2020_9d752cb0}
\bibfield{author}{\bibinfo{person}{Marwa El~Halabi}, \bibinfo{person}{Slobodan Mitrovi\'{c}}, \bibinfo{person}{Ashkan Norouzi-Fard}, \bibinfo{person}{Jakab Tardos}, {and} \bibinfo{person}{Jakub~M Tarnawski}.} \bibinfo{year}{2020}\natexlab{}.
\newblock \showarticletitle{Fairness in Streaming Submodular Maximization: Algorithms and Hardness}. In \bibinfo{booktitle}{\emph{Advances in Neural Information Processing Systems}}, \bibfield{editor}{\bibinfo{person}{H.~Larochelle}, \bibinfo{person}{M.~Ranzato}, \bibinfo{person}{R.~Hadsell}, \bibinfo{person}{M.F. Balcan}, {and} \bibinfo{person}{H.~Lin}} (Eds.), Vol.~\bibinfo{volume}{33}. \bibinfo{publisher}{Curran Associates, Inc.}, \bibinfo{pages}{13609--13622}.
\newblock
\urldef\tempurl%
\url{https://proceedings.neurips.cc/paper_files/paper/2020/file/9d752cb08ef466fc480fba981cfa44a1-Paper.pdf}
\showURL{%
\tempurl}


\bibitem[Ensign et~al\mbox{.}(2018)]%
        {pmlr-v81-ensign18a}
\bibfield{author}{\bibinfo{person}{Danielle Ensign}, \bibinfo{person}{Sorelle~A. Friedler}, \bibinfo{person}{Scott Neville}, \bibinfo{person}{Carlos Scheidegger}, {and} \bibinfo{person}{Suresh Venkatasubramanian}.} \bibinfo{year}{2018}\natexlab{}.
\newblock \showarticletitle{Runaway Feedback Loops in Predictive Policing}. In \bibinfo{booktitle}{\emph{Proceedings of the 1st Conference on Fairness, Accountability and Transparency}} \emph{(\bibinfo{series}{Proceedings of Machine Learning Research}, Vol.~\bibinfo{volume}{81})}, \bibfield{editor}{\bibinfo{person}{Sorelle~A. Friedler} {and} \bibinfo{person}{Christo Wilson}} (Eds.). \bibinfo{publisher}{PMLR}, \bibinfo{pages}{160--171}.
\newblock
\urldef\tempurl%
\url{https://proceedings.mlr.press/v81/ensign18a.html}
\showURL{%
\tempurl}


\bibitem[Farnadi et~al\mbox{.}(2018)]%
        {DBLP:journals/corr/abs-1809-09030}
\bibfield{author}{\bibinfo{person}{Golnoosh Farnadi}, \bibinfo{person}{Pigi Kouki}, \bibinfo{person}{Spencer~K. Thompson}, \bibinfo{person}{Sriram Srinivasan}, {and} \bibinfo{person}{Lise Getoor}.} \bibinfo{year}{2018}\natexlab{}.
\newblock \showarticletitle{A Fairness-aware Hybrid Recommender System}.
\newblock \bibinfo{journal}{\emph{CoRR}}  \bibinfo{volume}{abs/1809.09030} (\bibinfo{year}{2018}).
\newblock
\showeprint[arXiv]{1809.09030}
\urldef\tempurl%
\url{http://arxiv.org/abs/1809.09030}
\showURL{%
\tempurl}


\bibitem[Foulds et~al\mbox{.}(2019)]%
        {foulds2019bayesianmodelingintersectionalfairness}
\bibfield{author}{\bibinfo{person}{James Foulds}, \bibinfo{person}{Rashidul Islam}, \bibinfo{person}{Kamrun Keya}, {and} \bibinfo{person}{Shimei Pan}.} \bibinfo{year}{2019}\natexlab{}.
\newblock \bibinfo{title}{Bayesian Modeling of Intersectional Fairness: The Variance of Bias}.
\newblock
\showeprint[arxiv]{1811.07255}~[cs.LG]
\urldef\tempurl%
\url{https://arxiv.org/abs/1811.07255}
\showURL{%
\tempurl}


\bibitem[Foulds et~al\mbox{.}(2020)]%
        {9101635}
\bibfield{author}{\bibinfo{person}{James~R. Foulds}, \bibinfo{person}{Rashidul Islam}, \bibinfo{person}{Kamrun~Naher Keya}, {and} \bibinfo{person}{Shimei Pan}.} \bibinfo{year}{2020}\natexlab{}.
\newblock \showarticletitle{An Intersectional Definition of Fairness}. In \bibinfo{booktitle}{\emph{2020 IEEE 36th International Conference on Data Engineering (ICDE)}}. \bibinfo{pages}{1918--1921}.
\newblock
\href{https://doi.org/10.1109/ICDE48307.2020.00203}{doi:\nolinkurl{10.1109/ICDE48307.2020.00203}}


\bibitem[Fu et~al\mbox{.}(2020)]%
        {Fu20}
\bibfield{author}{\bibinfo{person}{Zuohui Fu}, \bibinfo{person}{Yikun Xian}, \bibinfo{person}{Ruoyuan Gao}, \bibinfo{person}{Jieyu Zhao}, \bibinfo{person}{Qiaoying Huang}, \bibinfo{person}{Yingqiang Ge}, \bibinfo{person}{Shuyuan Xu}, \bibinfo{person}{Shijie Geng}, \bibinfo{person}{Chirag Shah}, \bibinfo{person}{Yongfeng Zhang}, {and} \bibinfo{person}{Gerard de Melo}.} \bibinfo{year}{2020}\natexlab{}.
\newblock \showarticletitle{Fairness-Aware Explainable Recommendation over Knowledge Graphs}. In \bibinfo{booktitle}{\emph{Proceedings of the 43rd International ACM SIGIR Conference on Research and Development in Information Retrieval}} (Virtual Event, China) \emph{(\bibinfo{series}{SIGIR '20})}. \bibinfo{publisher}{Association for Computing Machinery}, \bibinfo{address}{New York, NY, USA}, \bibinfo{pages}{69–78}.
\newblock
\showISBNx{9781450380164}
\href{https://doi.org/10.1145/3397271.3401051}{doi:\nolinkurl{10.1145/3397271.3401051}}


\bibitem[Galhotra et~al\mbox{.}(2017)]%
        {10.1145/3106237.3106277}
\bibfield{author}{\bibinfo{person}{Sainyam Galhotra}, \bibinfo{person}{Yuriy Brun}, {and} \bibinfo{person}{Alexandra Meliou}.} \bibinfo{year}{2017}\natexlab{}.
\newblock \showarticletitle{Fairness testing: testing software for discrimination}. In \bibinfo{booktitle}{\emph{Proceedings of the 2017 11th Joint Meeting on Foundations of Software Engineering}} (Paderborn, Germany) \emph{(\bibinfo{series}{ESEC/FSE 2017})}. \bibinfo{publisher}{Association for Computing Machinery}, \bibinfo{address}{New York, NY, USA}, \bibinfo{pages}{498–510}.
\newblock
\showISBNx{9781450351058}
\href{https://doi.org/10.1145/3106237.3106277}{doi:\nolinkurl{10.1145/3106237.3106277}}


\bibitem[Ge et~al\mbox{.}(2021)]%
        {10.1145/3437963.3441824}
\bibfield{author}{\bibinfo{person}{Yingqiang Ge}, \bibinfo{person}{Shuchang Liu}, \bibinfo{person}{Ruoyuan Gao}, \bibinfo{person}{Yikun Xian}, \bibinfo{person}{Yunqi Li}, \bibinfo{person}{Xiangyu Zhao}, \bibinfo{person}{Changhua Pei}, \bibinfo{person}{Fei Sun}, \bibinfo{person}{Junfeng Ge}, \bibinfo{person}{Wenwu Ou}, {and} \bibinfo{person}{Yongfeng Zhang}.} \bibinfo{year}{2021}\natexlab{}.
\newblock \showarticletitle{Towards Long-term Fairness in Recommendation}. In \bibinfo{booktitle}{\emph{Proceedings of the 14th ACM International Conference on Web Search and Data Mining}} \emph{(\bibinfo{series}{WSDM '21})}. \bibinfo{publisher}{Association for Computing Machinery}, \bibinfo{address}{New York, NY, USA}, \bibinfo{pages}{445–453}.
\newblock
\showISBNx{9781450382977}
\href{https://doi.org/10.1145/3437963.3441824}{doi:\nolinkurl{10.1145/3437963.3441824}}


\bibitem[Geyik et~al\mbox{.}(2019)]%
        {10.1145/3292500.3330691}
\bibfield{author}{\bibinfo{person}{Sahin~Cem Geyik}, \bibinfo{person}{Stuart Ambler}, {and} \bibinfo{person}{Krishnaram Kenthapadi}.} \bibinfo{year}{2019}\natexlab{}.
\newblock \showarticletitle{Fairness-Aware Ranking in Search \& Recommendation Systems with Application to LinkedIn Talent Search}. In \bibinfo{booktitle}{\emph{Proceedings of the 25th ACM SIGKDD International Conference on Knowledge Discovery \& Data Mining}} (Anchorage, AK, USA) \emph{(\bibinfo{series}{KDD '19})}. \bibinfo{publisher}{Association for Computing Machinery}, \bibinfo{address}{New York, NY, USA}, \bibinfo{pages}{2221–2231}.
\newblock
\showISBNx{9781450362016}
\href{https://doi.org/10.1145/3292500.3330691}{doi:\nolinkurl{10.1145/3292500.3330691}}


\bibitem[Ghosh et~al\mbox{.}(2021)]%
        {ghosh21}
\bibfield{author}{\bibinfo{person}{Avijit Ghosh}, \bibinfo{person}{Ritam Dutt}, {and} \bibinfo{person}{Christo Wilson}.} \bibinfo{year}{2021}\natexlab{}.
\newblock \showarticletitle{When Fair Ranking Meets Uncertain Inference}. In \bibinfo{booktitle}{\emph{Proceedings of the 44th International ACM SIGIR Conference on Research and Development in Information Retrieval}} (Virtual Event, Canada) \emph{(\bibinfo{series}{SIGIR '21})}. \bibinfo{publisher}{Association for Computing Machinery}, \bibinfo{address}{New York, NY, USA}, \bibinfo{pages}{1033–1043}.
\newblock
\showISBNx{9781450380379}
\href{https://doi.org/10.1145/3404835.3462850}{doi:\nolinkurl{10.1145/3404835.3462850}}


\bibitem[Gohar and Cheng(2023)]%
        {Gohar2023ASO}
\bibfield{author}{\bibinfo{person}{Usman Gohar} {and} \bibinfo{person}{Lu Cheng}.} \bibinfo{year}{2023}\natexlab{}.
\newblock \showarticletitle{A Survey on Intersectional Fairness in Machine Learning: Notions, Mitigation, and Challenges}. In \bibinfo{booktitle}{\emph{International Joint Conference on Artificial Intelligence}}.
\newblock
\urldef\tempurl%
\url{https://api.semanticscholar.org/CorpusID:258615326}
\showURL{%
\tempurl}


\bibitem[Gorantla et~al\mbox{.}(2021)]%
        {pmlr-v139-gorantla21a}
\bibfield{author}{\bibinfo{person}{Sruthi Gorantla}, \bibinfo{person}{Amit Deshpande}, {and} \bibinfo{person}{Anand Louis}.} \bibinfo{year}{2021}\natexlab{}.
\newblock \showarticletitle{On the Problem of Underranking in Group-Fair Ranking}. In \bibinfo{booktitle}{\emph{Proceedings of the 38th International Conference on Machine Learning}} \emph{(\bibinfo{series}{Proceedings of Machine Learning Research}, Vol.~\bibinfo{volume}{139})}, \bibfield{editor}{\bibinfo{person}{Marina Meila} {and} \bibinfo{person}{Tong Zhang}} (Eds.). \bibinfo{publisher}{PMLR}, \bibinfo{pages}{3777--3787}.
\newblock
\urldef\tempurl%
\url{https://proceedings.mlr.press/v139/gorantla21a.html}
\showURL{%
\tempurl}


\bibitem[Gupta et~al\mbox{.}(2022)]%
        {gupta_questioning_2022}
\bibfield{author}{\bibinfo{person}{Manjul Gupta}, \bibinfo{person}{Carlos~M. Parra}, {and} \bibinfo{person}{Denis Dennehy}.} \bibinfo{year}{2022}\natexlab{}.
\newblock \showarticletitle{Questioning {Racial} and {Gender} {Bias} in {AI}-based {Recommendations}: {Do} {Espoused} {National} {Cultural} {Values} {Matter}?}
\newblock \bibinfo{journal}{\emph{Information Systems Frontiers}} \bibinfo{volume}{24}, \bibinfo{number}{5} (\bibinfo{date}{Oct.} \bibinfo{year}{2022}), \bibinfo{pages}{1465--1481}.
\newblock
\showISSN{1572-9419}
\href{https://doi.org/10.1007/s10796-021-10156-2}{doi:\nolinkurl{10.1007/s10796-021-10156-2}}


\bibitem[Hao et~al\mbox{.}(2021)]%
        {hao21}
\bibfield{author}{\bibinfo{person}{Qianxiu Hao}, \bibinfo{person}{Qianqian Xu}, \bibinfo{person}{Zhiyong Yang}, {and} \bibinfo{person}{Qingming Huang}.} \bibinfo{year}{2021}\natexlab{}.
\newblock \showarticletitle{Pareto Optimality for Fairness-constrained Collaborative Filtering}. In \bibinfo{booktitle}{\emph{Proceedings of the 29th ACM International Conference on Multimedia}} (Virtual Event, China) \emph{(\bibinfo{series}{MM '21})}. \bibinfo{publisher}{Association for Computing Machinery}, \bibinfo{address}{New York, NY, USA}, \bibinfo{pages}{5619–5627}.
\newblock
\showISBNx{9781450386517}
\href{https://doi.org/10.1145/3474085.3475706}{doi:\nolinkurl{10.1145/3474085.3475706}}


\bibitem[Hardt et~al\mbox{.}(2016)]%
        {hardt16}
\bibfield{author}{\bibinfo{person}{Moritz Hardt}, \bibinfo{person}{Eric Price}, {and} \bibinfo{person}{Nathan Srebro}.} \bibinfo{year}{2016}\natexlab{}.
\newblock \showarticletitle{Equality of opportunity in supervised learning}. In \bibinfo{booktitle}{\emph{Proceedings of the 30th International Conference on Neural Information Processing Systems}} (Barcelona, Spain) \emph{(\bibinfo{series}{NIPS'16})}. \bibinfo{publisher}{Curran Associates Inc.}, \bibinfo{address}{Red Hook, NY, USA}, \bibinfo{pages}{3323–3331}.
\newblock
\showISBNx{9781510838819}


\bibitem[Harper and Konstan(2015)]%
        {10.1145/2827872}
\bibfield{author}{\bibinfo{person}{F.~Maxwell Harper} {and} \bibinfo{person}{Joseph~A. Konstan}.} \bibinfo{year}{2015}\natexlab{}.
\newblock \showarticletitle{The MovieLens Datasets: History and Context}.
\newblock \bibinfo{journal}{\emph{ACM Trans. Interact. Intell. Syst.}} \bibinfo{volume}{5}, \bibinfo{number}{4}, Article \bibinfo{articleno}{19} (\bibinfo{date}{Dec.} \bibinfo{year}{2015}), \bibinfo{numpages}{19}~pages.
\newblock
\showISSN{2160-6455}
\href{https://doi.org/10.1145/2827872}{doi:\nolinkurl{10.1145/2827872}}


\bibitem[He et~al\mbox{.}(2017)]%
        {10.1145/3038912.3052569}
\bibfield{author}{\bibinfo{person}{Xiangnan He}, \bibinfo{person}{Lizi Liao}, \bibinfo{person}{Hanwang Zhang}, \bibinfo{person}{Liqiang Nie}, \bibinfo{person}{Xia Hu}, {and} \bibinfo{person}{Tat-Seng Chua}.} \bibinfo{year}{2017}\natexlab{}.
\newblock \showarticletitle{Neural Collaborative Filtering}. In \bibinfo{booktitle}{\emph{Proceedings of the 26th International Conference on World Wide Web}} (Perth, Australia) \emph{(\bibinfo{series}{WWW '17})}. \bibinfo{publisher}{International World Wide Web Conferences Steering Committee}, \bibinfo{address}{Republic and Canton of Geneva, CHE}, \bibinfo{pages}{173–182}.
\newblock
\showISBNx{9781450349130}
\href{https://doi.org/10.1145/3038912.3052569}{doi:\nolinkurl{10.1145/3038912.3052569}}


\bibitem[Imana et~al\mbox{.}(2021)]%
        {10.1145/3442381.3450077}
\bibfield{author}{\bibinfo{person}{Basileal Imana}, \bibinfo{person}{Aleksandra Korolova}, {and} \bibinfo{person}{John Heidemann}.} \bibinfo{year}{2021}\natexlab{}.
\newblock \showarticletitle{Auditing for Discrimination in Algorithms Delivering Job Ads}. In \bibinfo{booktitle}{\emph{Proceedings of the Web Conference 2021}} (Ljubljana, Slovenia) \emph{(\bibinfo{series}{WWW '21})}. \bibinfo{publisher}{Association for Computing Machinery}, \bibinfo{address}{New York, NY, USA}, \bibinfo{pages}{3767–3778}.
\newblock
\showISBNx{9781450383127}
\href{https://doi.org/10.1145/3442381.3450077}{doi:\nolinkurl{10.1145/3442381.3450077}}


\bibitem[Islam et~al\mbox{.}(2021a)]%
        {islam21}
\bibfield{author}{\bibinfo{person}{Rashidul Islam}, \bibinfo{person}{Kamrun~Naher Keya}, \bibinfo{person}{Ziqian Zeng}, \bibinfo{person}{Shimei Pan}, {and} \bibinfo{person}{James Foulds}.} \bibinfo{year}{2021}\natexlab{a}.
\newblock \showarticletitle{Debiasing Career Recommendations with Neural Fair Collaborative Filtering}. In \bibinfo{booktitle}{\emph{Proceedings of the Web Conference 2021}} (Ljubljana, Slovenia) \emph{(\bibinfo{series}{WWW '21})}. \bibinfo{publisher}{Association for Computing Machinery}, \bibinfo{address}{New York, NY, USA}, \bibinfo{pages}{3779–3790}.
\newblock
\showISBNx{9781450383127}
\href{https://doi.org/10.1145/3442381.3449904}{doi:\nolinkurl{10.1145/3442381.3449904}}


\bibitem[Islam et~al\mbox{.}(2021b)]%
        {10.1145/3442381.3449904}
\bibfield{author}{\bibinfo{person}{Rashidul Islam}, \bibinfo{person}{Kamrun~Naher Keya}, \bibinfo{person}{Ziqian Zeng}, \bibinfo{person}{Shimei Pan}, {and} \bibinfo{person}{James Foulds}.} \bibinfo{year}{2021}\natexlab{b}.
\newblock \showarticletitle{Debiasing Career Recommendations with Neural Fair Collaborative Filtering}. In \bibinfo{booktitle}{\emph{Proceedings of the Web Conference 2021}} (Ljubljana, Slovenia) \emph{(\bibinfo{series}{WWW '21})}. \bibinfo{publisher}{Association for Computing Machinery}, \bibinfo{address}{New York, NY, USA}, \bibinfo{pages}{3779–3790}.
\newblock
\showISBNx{9781450383127}
\href{https://doi.org/10.1145/3442381.3449904}{doi:\nolinkurl{10.1145/3442381.3449904}}


\bibitem[Jaenich et~al\mbox{.}(2024)]%
        {10.1145/3626772.3657794}
\bibfield{author}{\bibinfo{person}{Thomas Jaenich}, \bibinfo{person}{Graham McDonald}, {and} \bibinfo{person}{Iadh Ounis}.} \bibinfo{year}{2024}\natexlab{}.
\newblock \showarticletitle{Fairness-Aware Exposure Allocation via Adaptive Reranking}. In \bibinfo{booktitle}{\emph{Proceedings of the 47th International ACM SIGIR Conference on Research and Development in Information Retrieval}} (Washington DC, USA) \emph{(\bibinfo{series}{SIGIR '24})}. \bibinfo{publisher}{Association for Computing Machinery}, \bibinfo{address}{New York, NY, USA}, \bibinfo{pages}{1504–1513}.
\newblock
\showISBNx{9798400704314}
\href{https://doi.org/10.1145/3626772.3657794}{doi:\nolinkurl{10.1145/3626772.3657794}}


\bibitem[Jiang et~al\mbox{.}(2024)]%
        {10.1145/3589334.3648158}
\bibfield{author}{\bibinfo{person}{Meng Jiang}, \bibinfo{person}{Keqin Bao}, \bibinfo{person}{Jizhi Zhang}, \bibinfo{person}{Wenjie Wang}, \bibinfo{person}{Zhengyi Yang}, \bibinfo{person}{Fuli Feng}, {and} \bibinfo{person}{Xiangnan He}.} \bibinfo{year}{2024}\natexlab{}.
\newblock \showarticletitle{Item-side Fairness of Large Language Model-based Recommendation System}. In \bibinfo{booktitle}{\emph{Proceedings of the ACM Web Conference 2024}} (Singapore, Singapore) \emph{(\bibinfo{series}{WWW '24})}. \bibinfo{publisher}{Association for Computing Machinery}, \bibinfo{address}{New York, NY, USA}, \bibinfo{pages}{4717–4726}.
\newblock
\showISBNx{9798400701719}
\href{https://doi.org/10.1145/3589334.3648158}{doi:\nolinkurl{10.1145/3589334.3648158}}


\bibitem[Jin et~al\mbox{.}(2023)]%
        {JIN2023101906}
\bibfield{author}{\bibinfo{person}{Di Jin}, \bibinfo{person}{Luzhi Wang}, \bibinfo{person}{He Zhang}, \bibinfo{person}{Yizhen Zheng}, \bibinfo{person}{Weiping Ding}, \bibinfo{person}{Feng Xia}, {and} \bibinfo{person}{Shirui Pan}.} \bibinfo{year}{2023}\natexlab{}.
\newblock \showarticletitle{A survey on fairness-aware recommender systems}.
\newblock \bibinfo{journal}{\emph{Information Fusion}}  \bibinfo{volume}{100} (\bibinfo{year}{2023}), \bibinfo{pages}{101906}.
\newblock
\showISSN{1566-2535}
\href{https://doi.org/10.1016/j.inffus.2023.101906}{doi:\nolinkurl{10.1016/j.inffus.2023.101906}}


\bibitem[Keya et~al\mbox{.}(2020)]%
        {keya2020equitableallocationhealthcareresources}
\bibfield{author}{\bibinfo{person}{Kamrun~Naher Keya}, \bibinfo{person}{Rashidul Islam}, \bibinfo{person}{Shimei Pan}, \bibinfo{person}{Ian Stockwell}, {and} \bibinfo{person}{James~R. Foulds}.} \bibinfo{year}{2020}\natexlab{}.
\newblock \bibinfo{title}{Equitable Allocation of Healthcare Resources with Fair Cox Models}.
\newblock
\showeprint[arxiv]{2010.06820}~[cs.LG]
\urldef\tempurl%
\url{https://arxiv.org/abs/2010.06820}
\showURL{%
\tempurl}


\bibitem[Kheya et~al\mbox{.}(2024)]%
        {kheya2024pursuitfairnessartificialintelligence}
\bibfield{author}{\bibinfo{person}{Tahsin~Alamgir Kheya}, \bibinfo{person}{Mohamed~Reda Bouadjenek}, {and} \bibinfo{person}{Sunil Aryal}.} \bibinfo{year}{2024}\natexlab{}.
\newblock \bibinfo{title}{The Pursuit of Fairness in Artificial Intelligence Models: A Survey}.
\newblock
\showeprint[arxiv]{2403.17333}~[cs.AI]
\urldef\tempurl%
\url{https://arxiv.org/abs/2403.17333}
\showURL{%
\tempurl}


\bibitem[Koren et~al\mbox{.}(2009)]%
        {5197422}
\bibfield{author}{\bibinfo{person}{Yehuda Koren}, \bibinfo{person}{Robert Bell}, {and} \bibinfo{person}{Chris Volinsky}.} \bibinfo{year}{2009}\natexlab{}.
\newblock \showarticletitle{Matrix Factorization Techniques for Recommender Systems}.
\newblock \bibinfo{journal}{\emph{Computer}} \bibinfo{volume}{42}, \bibinfo{number}{8} (\bibinfo{year}{2009}), \bibinfo{pages}{30--37}.
\newblock
\href{https://doi.org/10.1109/MC.2009.263}{doi:\nolinkurl{10.1109/MC.2009.263}}


\bibitem[Kusner et~al\mbox{.}(2017)]%
        {10.5555/3294996.3295162}
\bibfield{author}{\bibinfo{person}{Matt Kusner}, \bibinfo{person}{Joshua Loftus}, \bibinfo{person}{Chris Russell}, {and} \bibinfo{person}{Ricardo Silva}.} \bibinfo{year}{2017}\natexlab{}.
\newblock \showarticletitle{Counterfactual fairness}. In \bibinfo{booktitle}{\emph{Proceedings of the 31st International Conference on Neural Information Processing Systems}} (Long Beach, California, USA) \emph{(\bibinfo{series}{NIPS'17})}. \bibinfo{publisher}{Curran Associates Inc.}, \bibinfo{address}{Red Hook, NY, USA}, \bibinfo{pages}{4069–4079}.
\newblock
\showISBNx{9781510860964}


\bibitem[Li et~al\mbox{.}(2021a)]%
        {10.1145/3442381.3449866}
\bibfield{author}{\bibinfo{person}{Yunqi Li}, \bibinfo{person}{Hanxiong Chen}, \bibinfo{person}{Zuohui Fu}, \bibinfo{person}{Yingqiang Ge}, {and} \bibinfo{person}{Yongfeng Zhang}.} \bibinfo{year}{2021}\natexlab{a}.
\newblock \showarticletitle{User-oriented Fairness in Recommendation}. In \bibinfo{booktitle}{\emph{Proceedings of the Web Conference 2021}} (Ljubljana, Slovenia) \emph{(\bibinfo{series}{WWW '21})}. \bibinfo{publisher}{Association for Computing Machinery}, \bibinfo{address}{New York, NY, USA}, \bibinfo{pages}{624–632}.
\newblock
\showISBNx{9781450383127}
\href{https://doi.org/10.1145/3442381.3449866}{doi:\nolinkurl{10.1145/3442381.3449866}}


\bibitem[Li et~al\mbox{.}(2021b)]%
        {li2021towards}
\bibfield{author}{\bibinfo{person}{Yunqi Li}, \bibinfo{person}{Hanxiong Chen}, \bibinfo{person}{Shuyuan Xu}, \bibinfo{person}{Yingqiang Ge}, {and} \bibinfo{person}{Yongfeng Zhang}.} \bibinfo{year}{2021}\natexlab{b}.
\newblock \showarticletitle{Towards personalized fairness based on causal notion}. In \bibinfo{booktitle}{\emph{Proceedings of the 44th International ACM SIGIR Conference on Research and Development in Information Retrieval}}. \bibinfo{pages}{1054--1063}.
\newblock


\bibitem[Liang et~al\mbox{.}(2018)]%
        {10.1145/3178876.3186150}
\bibfield{author}{\bibinfo{person}{Dawen Liang}, \bibinfo{person}{Rahul~G. Krishnan}, \bibinfo{person}{Matthew~D. Hoffman}, {and} \bibinfo{person}{Tony Jebara}.} \bibinfo{year}{2018}\natexlab{}.
\newblock \showarticletitle{Variational Autoencoders for Collaborative Filtering}. In \bibinfo{booktitle}{\emph{Proceedings of the 2018 World Wide Web Conference}} (Lyon, France) \emph{(\bibinfo{series}{WWW '18})}. \bibinfo{publisher}{International World Wide Web Conferences Steering Committee}, \bibinfo{address}{Republic and Canton of Geneva, CHE}, \bibinfo{pages}{689–698}.
\newblock
\showISBNx{9781450356398}
\href{https://doi.org/10.1145/3178876.3186150}{doi:\nolinkurl{10.1145/3178876.3186150}}


\bibitem[Lin et~al\mbox{.}(2021)]%
        {10.1145/3404835.3462943}
\bibfield{author}{\bibinfo{person}{Chen Lin}, \bibinfo{person}{Xinyi Liu}, \bibinfo{person}{Guipeng Xv}, {and} \bibinfo{person}{Hui Li}.} \bibinfo{year}{2021}\natexlab{}.
\newblock \showarticletitle{Mitigating Sentiment Bias for Recommender Systems}. In \bibinfo{booktitle}{\emph{Proceedings of the 44th International ACM SIGIR Conference on Research and Development in Information Retrieval}} (Virtual Event, Canada) \emph{(\bibinfo{series}{SIGIR '21})}. \bibinfo{publisher}{Association for Computing Machinery}, \bibinfo{address}{New York, NY, USA}, \bibinfo{pages}{31–40}.
\newblock
\showISBNx{9781450380379}
\href{https://doi.org/10.1145/3404835.3462943}{doi:\nolinkurl{10.1145/3404835.3462943}}


\bibitem[Lin et~al\mbox{.}(2019)]%
        {lin2019crankvolumepreferencebias}
\bibfield{author}{\bibinfo{person}{Kun Lin}, \bibinfo{person}{Nasim Sonboli}, \bibinfo{person}{Bamshad Mobasher}, {and} \bibinfo{person}{Robin Burke}.} \bibinfo{year}{2019}\natexlab{}.
\newblock \bibinfo{title}{Crank up the volume: preference bias amplification in collaborative recommendation}.
\newblock
\showeprint[arxiv]{1909.06362}~[cs.IR]
\urldef\tempurl%
\url{https://arxiv.org/abs/1909.06362}
\showURL{%
\tempurl}


\bibitem[Liu et~al\mbox{.}(2022)]%
        {LIU2022108058}
\bibfield{author}{\bibinfo{person}{Haifeng Liu}, \bibinfo{person}{Nan Zhao}, \bibinfo{person}{Xiaokun Zhang}, \bibinfo{person}{Hongfei Lin}, \bibinfo{person}{Liang Yang}, \bibinfo{person}{Bo Xu}, \bibinfo{person}{Yuan Lin}, {and} \bibinfo{person}{Wenqi Fan}.} \bibinfo{year}{2022}\natexlab{}.
\newblock \showarticletitle{Dual constraints and adversarial learning for fair recommenders}.
\newblock \bibinfo{journal}{\emph{Knowledge-Based Systems}}  \bibinfo{volume}{239} (\bibinfo{year}{2022}), \bibinfo{pages}{108058}.
\newblock
\showISSN{0950-7051}
\href{https://doi.org/10.1016/j.knosys.2021.108058}{doi:\nolinkurl{10.1016/j.knosys.2021.108058}}


\bibitem[Lum and Isaac(2016)]%
        {lum2016}
\bibfield{author}{\bibinfo{person}{Kristian Lum} {and} \bibinfo{person}{William Isaac}.} \bibinfo{year}{2016}\natexlab{}.
\newblock \showarticletitle{To Predict and Serve?}
\newblock \bibinfo{journal}{\emph{Significance}}  \bibinfo{volume}{13} (\bibinfo{date}{10} \bibinfo{year}{2016}), \bibinfo{pages}{14--19}.
\newblock
\href{https://doi.org/10.1111/j.1740-9713.2016.00960.x}{doi:\nolinkurl{10.1111/j.1740-9713.2016.00960.x}}


\bibitem[Mansoury et~al\mbox{.}(2019)]%
        {mansoury2019biasdisparitycollaborativerecommendation}
\bibfield{author}{\bibinfo{person}{Masoud Mansoury}, \bibinfo{person}{Bamshad Mobasher}, \bibinfo{person}{Robin Burke}, {and} \bibinfo{person}{Mykola Pechenizkiy}.} \bibinfo{year}{2019}\natexlab{}.
\newblock \bibinfo{title}{Bias Disparity in Collaborative Recommendation: Algorithmic Evaluation and Comparison}.
\newblock
\showeprint[arxiv]{1908.00831}~[cs.IR]
\urldef\tempurl%
\url{https://arxiv.org/abs/1908.00831}
\showURL{%
\tempurl}


\bibitem[Medda et~al\mbox{.}(2024)]%
        {10.1145/3655631}
\bibfield{author}{\bibinfo{person}{Giacomo Medda}, \bibinfo{person}{Francesco Fabbri}, \bibinfo{person}{Mirko Marras}, \bibinfo{person}{Ludovico Boratto}, {and} \bibinfo{person}{Gianni Fenu}.} \bibinfo{year}{2024}\natexlab{}.
\newblock \showarticletitle{GNNUERS: Fairness Explanation in GNNs for Recommendation via Counterfactual Reasoning}.
\newblock \bibinfo{journal}{\emph{ACM Trans. Intell. Syst. Technol.}} (\bibinfo{date}{apr} \bibinfo{year}{2024}).
\newblock
\showISSN{2157-6904}
\href{https://doi.org/10.1145/3655631}{doi:\nolinkurl{10.1145/3655631}}
\newblock
\shownote{Just Accepted}.


\bibitem[Melchiorre et~al\mbox{.}(2021)]%
        {Melchiorre21}
\bibfield{author}{\bibinfo{person}{Alessandro~B. Melchiorre}, \bibinfo{person}{Navid Rekabsaz}, \bibinfo{person}{Emilia Parada-Cabaleiro}, \bibinfo{person}{Stefan Brandl}, \bibinfo{person}{Oleg Lesota}, {and} \bibinfo{person}{Markus Schedl}.} \bibinfo{year}{2021}\natexlab{}.
\newblock \showarticletitle{Investigating gender fairness of recommendation algorithms in the music domain}.
\newblock \bibinfo{journal}{\emph{Inf. Process. Manage.}} \bibinfo{volume}{58}, \bibinfo{number}{5} (\bibinfo{date}{sep} \bibinfo{year}{2021}), \bibinfo{numpages}{27}~pages.
\newblock
\showISSN{0306-4573}
\href{https://doi.org/10.1016/j.ipm.2021.102666}{doi:\nolinkurl{10.1016/j.ipm.2021.102666}}


\bibitem[Naghiaei et~al\mbox{.}(2022)]%
        {10.1145/3477495.3531959}
\bibfield{author}{\bibinfo{person}{Mohammadmehdi Naghiaei}, \bibinfo{person}{Hossein~A. Rahmani}, {and} \bibinfo{person}{Yashar Deldjoo}.} \bibinfo{year}{2022}\natexlab{}.
\newblock \showarticletitle{CPFair: Personalized Consumer and Producer Fairness Re-ranking for Recommender Systems}. In \bibinfo{booktitle}{\emph{Proceedings of the 45th International ACM SIGIR Conference on Research and Development in Information Retrieval}} (Madrid, Spain) \emph{(\bibinfo{series}{SIGIR '22})}. \bibinfo{publisher}{Association for Computing Machinery}, \bibinfo{address}{New York, NY, USA}, \bibinfo{pages}{770–779}.
\newblock
\showISBNx{9781450387323}
\href{https://doi.org/10.1145/3477495.3531959}{doi:\nolinkurl{10.1145/3477495.3531959}}


\bibitem[Rahmani et~al\mbox{.}(2024)]%
        {10.1145/3651167}
\bibfield{author}{\bibinfo{person}{Hossein~A. Rahmani}, \bibinfo{person}{Mohammadmehdi Naghiaei}, {and} \bibinfo{person}{Yashar Deldjoo}.} \bibinfo{year}{2024}\natexlab{}.
\newblock \showarticletitle{A Personalized Framework for Consumer and Producer Group Fairness Optimization in Recommender Systems}.
\newblock \bibinfo{journal}{\emph{ACM Trans. Recomm. Syst.}} \bibinfo{volume}{2}, \bibinfo{number}{3}, Article \bibinfo{articleno}{19} (\bibinfo{date}{jun} \bibinfo{year}{2024}), \bibinfo{numpages}{24}~pages.
\newblock
\href{https://doi.org/10.1145/3651167}{doi:\nolinkurl{10.1145/3651167}}


\bibitem[Rendle et~al\mbox{.}(2009)]%
        {10.5555/1795114.1795167}
\bibfield{author}{\bibinfo{person}{Steffen Rendle}, \bibinfo{person}{Christoph Freudenthaler}, \bibinfo{person}{Zeno Gantner}, {and} \bibinfo{person}{Lars Schmidt-Thieme}.} \bibinfo{year}{2009}\natexlab{}.
\newblock \showarticletitle{BPR: Bayesian personalized ranking from implicit feedback}. In \bibinfo{booktitle}{\emph{Proceedings of the Twenty-Fifth Conference on Uncertainty in Artificial Intelligence}} (Montreal, Quebec, Canada) \emph{(\bibinfo{series}{UAI '09})}. \bibinfo{publisher}{AUAI Press}, \bibinfo{address}{Arlington, Virginia, USA}, \bibinfo{pages}{452–461}.
\newblock
\showISBNx{9780974903958}


\bibitem[Rus et~al\mbox{.}(2022)]%
        {rus2022closinggenderwagegap}
\bibfield{author}{\bibinfo{person}{Clara Rus}, \bibinfo{person}{Jeffrey Luppes}, \bibinfo{person}{Harrie Oosterhuis}, {and} \bibinfo{person}{Gido~H. Schoenmacker}.} \bibinfo{year}{2022}\natexlab{}.
\newblock \bibinfo{title}{Closing the Gender Wage Gap: Adversarial Fairness in Job Recommendation}.
\newblock
\showeprint[arxiv]{2209.09592}~[cs.LG]
\urldef\tempurl%
\url{https://arxiv.org/abs/2209.09592}
\showURL{%
\tempurl}


\bibitem[Salah et~al\mbox{.}(2020)]%
        {JMLR:v21:19-805}
\bibfield{author}{\bibinfo{person}{Aghiles Salah}, \bibinfo{person}{Quoc-Tuan Truong}, {and} \bibinfo{person}{Hady~W. Lauw}.} \bibinfo{year}{2020}\natexlab{}.
\newblock \showarticletitle{Cornac: A Comparative Framework for Multimodal Recommender Systems}.
\newblock \bibinfo{journal}{\emph{Journal of Machine Learning Research}} \bibinfo{volume}{21}, \bibinfo{number}{95} (\bibinfo{year}{2020}), \bibinfo{pages}{1--5}.
\newblock
\urldef\tempurl%
\url{http://jmlr.org/papers/v21/19-805.html}
\showURL{%
\tempurl}


\bibitem[Sapiezynski et~al\mbox{.}(2019)]%
        {10.1145/3308560.3317595}
\bibfield{author}{\bibinfo{person}{Piotr Sapiezynski}, \bibinfo{person}{Wesley Zeng}, \bibinfo{person}{Ronald E~Robertson}, \bibinfo{person}{Alan Mislove}, {and} \bibinfo{person}{Christo Wilson}.} \bibinfo{year}{2019}\natexlab{}.
\newblock \showarticletitle{Quantifying the Impact of User Attentionon Fair Group Representation in Ranked Lists}. In \bibinfo{booktitle}{\emph{Companion Proceedings of The 2019 World Wide Web Conference}} (San Francisco, USA) \emph{(\bibinfo{series}{WWW '19})}. \bibinfo{publisher}{Association for Computing Machinery}, \bibinfo{address}{New York, NY, USA}, \bibinfo{pages}{553–562}.
\newblock
\showISBNx{9781450366755}
\href{https://doi.org/10.1145/3308560.3317595}{doi:\nolinkurl{10.1145/3308560.3317595}}


\bibitem[Sarwar et~al\mbox{.}(2000)]%
        {10.1145/352871.352887}
\bibfield{author}{\bibinfo{person}{Badrul Sarwar}, \bibinfo{person}{George Karypis}, \bibinfo{person}{Joseph Konstan}, {and} \bibinfo{person}{John Riedl}.} \bibinfo{year}{2000}\natexlab{}.
\newblock \showarticletitle{Analysis of recommendation algorithms for e-commerce}. In \bibinfo{booktitle}{\emph{Proceedings of the 2nd ACM Conference on Electronic Commerce}} (Minneapolis, Minnesota, USA) \emph{(\bibinfo{series}{EC '00})}. \bibinfo{publisher}{Association for Computing Machinery}, \bibinfo{address}{New York, NY, USA}, \bibinfo{pages}{158–167}.
\newblock
\showISBNx{1581132727}
\href{https://doi.org/10.1145/352871.352887}{doi:\nolinkurl{10.1145/352871.352887}}


\bibitem[Serbos et~al\mbox{.}(2017)]%
        {10.1145/3038912.3052612}
\bibfield{author}{\bibinfo{person}{Dimitris Serbos}, \bibinfo{person}{Shuyao Qi}, \bibinfo{person}{Nikos Mamoulis}, \bibinfo{person}{Evaggelia Pitoura}, {and} \bibinfo{person}{Panayiotis Tsaparas}.} \bibinfo{year}{2017}\natexlab{}.
\newblock \showarticletitle{Fairness in Package-to-Group Recommendations}. In \bibinfo{booktitle}{\emph{Proceedings of the 26th International Conference on World Wide Web}} (Perth, Australia) \emph{(\bibinfo{series}{WWW '17})}. \bibinfo{publisher}{International World Wide Web Conferences Steering Committee}, \bibinfo{address}{Republic and Canton of Geneva, CHE}, \bibinfo{pages}{371–379}.
\newblock
\showISBNx{9781450349130}
\href{https://doi.org/10.1145/3038912.3052612}{doi:\nolinkurl{10.1145/3038912.3052612}}


\bibitem[Shani and Gunawardana(2011)]%
        {Shani2021}
\bibfield{author}{\bibinfo{person}{Guy Shani} {and} \bibinfo{person}{Asela Gunawardana}.} \bibinfo{year}{2011}\natexlab{}.
\newblock \bibinfo{booktitle}{\emph{Evaluating Recommendation Systems}}. Vol.~\bibinfo{volume}{12}.
\newblock \bibinfo{pages}{257--297}.
\newblock
\showISBNx{978-0-387-85819-7}
\href{https://doi.org/10.1007/978-0-387-85820-3_8}{doi:\nolinkurl{10.1007/978-0-387-85820-3_8}}


\bibitem[Steck(2018)]%
        {10.1145/3240323.3240372}
\bibfield{author}{\bibinfo{person}{Harald Steck}.} \bibinfo{year}{2018}\natexlab{}.
\newblock \showarticletitle{Calibrated recommendations}. In \bibinfo{booktitle}{\emph{Proceedings of the 12th ACM Conference on Recommender Systems}} (Vancouver, British Columbia, Canada) \emph{(\bibinfo{series}{RecSys '18})}. \bibinfo{publisher}{Association for Computing Machinery}, \bibinfo{address}{New York, NY, USA}, \bibinfo{pages}{154–162}.
\newblock
\showISBNx{9781450359016}
\href{https://doi.org/10.1145/3240323.3240372}{doi:\nolinkurl{10.1145/3240323.3240372}}


\bibitem[Tang et~al\mbox{.}(2023)]%
        {tang_when_2023}
\bibfield{author}{\bibinfo{person}{Jiakai Tang}, \bibinfo{person}{Shiqi Shen}, \bibinfo{person}{Zhipeng Wang}, \bibinfo{person}{Zhi Gong}, \bibinfo{person}{Jingsen Zhang}, {and} \bibinfo{person}{Xu Chen}.} \bibinfo{year}{2023}\natexlab{}.
\newblock \showarticletitle{When {Fairness} meets {Bias}: a {Debiased} {Framework} for {Fairness} aware {Top}-{N} {Recommendation}}. In \bibinfo{booktitle}{\emph{Proceedings of the 17th {ACM} {Conference} on {Recommender} {Systems}}} \emph{(\bibinfo{series}{{RecSys} '23})}. \bibinfo{publisher}{Association for Computing Machinery}, \bibinfo{address}{New York, NY, USA}, \bibinfo{pages}{200--210}.
\newblock
\showISBNx{9798400702419}
\href{https://doi.org/10.1145/3604915.3608770}{doi:\nolinkurl{10.1145/3604915.3608770}}


\bibitem[Tsintzou et~al\mbox{.}(2019)]%
        {TsintzouPT19}
\bibfield{author}{\bibinfo{person}{Virginia Tsintzou}, \bibinfo{person}{Evaggelia Pitoura}, {and} \bibinfo{person}{Panayiotis Tsaparas}.} \bibinfo{year}{2019}\natexlab{}.
\newblock \showarticletitle{Bias Disparity in Recommendation Systems}. In \bibinfo{booktitle}{\emph{Proceedings of the Workshop on Recommendation in Multi-stakeholder Environments co-located with the 13th {ACM} Conference on Recommender Systems (RecSys 2019), Copenhagen, Denmark, September 20, 2019}} \emph{(\bibinfo{series}{{CEUR} Workshop Proceedings}, Vol.~\bibinfo{volume}{2440})}, \bibfield{editor}{\bibinfo{person}{Robin Burke}, \bibinfo{person}{Himan Abdollahpouri}, \bibinfo{person}{Edward~C. Malthouse}, \bibinfo{person}{K.~P. Thai}, {and} \bibinfo{person}{Yongfeng Zhang}} (Eds.). \bibinfo{publisher}{CEUR-WS.org}.
\newblock
\urldef\tempurl%
\url{https://ceur-ws.org/Vol-2440/short4.pdf}
\showURL{%
\tempurl}


\bibitem[Verma and Rubin(2018)]%
        {10.1145/3194770.3194776}
\bibfield{author}{\bibinfo{person}{Sahil Verma} {and} \bibinfo{person}{Julia Rubin}.} \bibinfo{year}{2018}\natexlab{}.
\newblock \showarticletitle{Fairness definitions explained}. In \bibinfo{booktitle}{\emph{Proceedings of the International Workshop on Software Fairness}} (Gothenburg, Sweden) \emph{(\bibinfo{series}{FairWare '18})}. \bibinfo{publisher}{Association for Computing Machinery}, \bibinfo{address}{New York, NY, USA}, \bibinfo{pages}{1–7}.
\newblock
\showISBNx{9781450357463}
\href{https://doi.org/10.1145/3194770.3194776}{doi:\nolinkurl{10.1145/3194770.3194776}}


\bibitem[Wan et~al\mbox{.}(2020)]%
        {wan20}
\bibfield{author}{\bibinfo{person}{Mengting Wan}, \bibinfo{person}{Jianmo Ni}, \bibinfo{person}{Rishabh Misra}, {and} \bibinfo{person}{Julian McAuley}.} \bibinfo{year}{2020}\natexlab{}.
\newblock \showarticletitle{Addressing Marketing Bias in Product Recommendations}. In \bibinfo{booktitle}{\emph{Proceedings of the 13th International Conference on Web Search and Data Mining}} (Houston, TX, USA) \emph{(\bibinfo{series}{WSDM '20})}. \bibinfo{publisher}{Association for Computing Machinery}, \bibinfo{address}{New York, NY, USA}, \bibinfo{pages}{618–626}.
\newblock
\showISBNx{9781450368223}
\href{https://doi.org/10.1145/3336191.3371855}{doi:\nolinkurl{10.1145/3336191.3371855}}


\bibitem[Wang et~al\mbox{.}(2023)]%
        {10.1145/3547333}
\bibfield{author}{\bibinfo{person}{Yifan Wang}, \bibinfo{person}{Weizhi Ma}, \bibinfo{person}{Min Zhang}, \bibinfo{person}{Yiqun Liu}, {and} \bibinfo{person}{Shaoping Ma}.} \bibinfo{year}{2023}\natexlab{}.
\newblock \showarticletitle{A Survey on the Fairness of Recommender Systems}.
\newblock \bibinfo{journal}{\emph{ACM Trans. Inf. Syst.}} \bibinfo{volume}{41}, \bibinfo{number}{3}, Article \bibinfo{articleno}{52} (\bibinfo{date}{feb} \bibinfo{year}{2023}), \bibinfo{numpages}{43}~pages.
\newblock
\showISSN{1046-8188}
\href{https://doi.org/10.1145/3547333}{doi:\nolinkurl{10.1145/3547333}}


\bibitem[Wei and He(2022a)]%
        {wei2022comprehensive}
\bibfield{author}{\bibinfo{person}{Tianxin Wei} {and} \bibinfo{person}{Jingrui He}.} \bibinfo{year}{2022}\natexlab{a}.
\newblock \showarticletitle{Comprehensive Fair Meta-learned Recommender System}. In \bibinfo{booktitle}{\emph{Proceedings of the 28th ACM SIGKDD Conference on Knowledge Discovery and Data Mining}}. \bibinfo{publisher}{Association for Computing Machinery}, \bibinfo{address}{New York, NY, USA}, \bibinfo{pages}{1989--1999}.
\newblock


\bibitem[Wei and He(2022b)]%
        {10.1145/3534678.3539269}
\bibfield{author}{\bibinfo{person}{Tianxin Wei} {and} \bibinfo{person}{Jingrui He}.} \bibinfo{year}{2022}\natexlab{b}.
\newblock \showarticletitle{Comprehensive Fair Meta-learned Recommender System}. In \bibinfo{booktitle}{\emph{Proceedings of the 28th ACM SIGKDD Conference on Knowledge Discovery and Data Mining}} (Washington DC, USA) \emph{(\bibinfo{series}{KDD '22})}. \bibinfo{publisher}{Association for Computing Machinery}, \bibinfo{address}{New York, NY, USA}, \bibinfo{pages}{1989–1999}.
\newblock
\showISBNx{9781450393850}
\href{https://doi.org/10.1145/3534678.3539269}{doi:\nolinkurl{10.1145/3534678.3539269}}


\bibitem[Weydemann et~al\mbox{.}(2019)]%
        {Weydemann19}
\bibfield{author}{\bibinfo{person}{Leonard Weydemann}, \bibinfo{person}{Dimitris Sacharidis}, {and} \bibinfo{person}{Hannes Werthner}.} \bibinfo{year}{2019}\natexlab{}.
\newblock \showarticletitle{Defining and measuring fairness in location recommendations}. In \bibinfo{booktitle}{\emph{Proceedings of the 3rd ACM SIGSPATIAL International Workshop on Location-Based Recommendations, Geosocial Networks and Geoadvertising}} (Chicago, Illinois) \emph{(\bibinfo{series}{LocalRec '19})}. \bibinfo{publisher}{Association for Computing Machinery}, \bibinfo{address}{New York, NY, USA}, Article \bibinfo{articleno}{6}, \bibinfo{numpages}{8}~pages.
\newblock
\showISBNx{9781450369633}
\href{https://doi.org/10.1145/3356994.3365497}{doi:\nolinkurl{10.1145/3356994.3365497}}


\bibitem[Wu et~al\mbox{.}(2020)]%
        {wu20}
\bibfield{author}{\bibinfo{person}{Chuhan Wu}, \bibinfo{person}{Fangzhao Wu}, \bibinfo{person}{Xiting Wang}, \bibinfo{person}{Yongfeng Huang}, {and} \bibinfo{person}{Xing Xie}.} \bibinfo{year}{2020}\natexlab{}.
\newblock \bibinfo{title}{Fairness-aware News Recommendation with Decomposed Adversarial Learning}.
\newblock


\bibitem[Wu et~al\mbox{.}(2022)]%
        {10068703}
\bibfield{author}{\bibinfo{person}{Kun Wu}, \bibinfo{person}{Jacob Erickson}, \bibinfo{person}{Wendy~Hui Wang}, {and} \bibinfo{person}{Yue Ning}.} \bibinfo{year}{2022}\natexlab{}.
\newblock \showarticletitle{Equipping Recommender Systems with Individual Fairness via Second-order Proximity Embedding}. In \bibinfo{booktitle}{\emph{2022 IEEE/ACM International Conference on Advances in Social Networks Analysis and Mining (ASONAM)}}. \bibinfo{pages}{171--175}.
\newblock
\href{https://doi.org/10.1109/ASONAM55673.2022.10068703}{doi:\nolinkurl{10.1109/ASONAM55673.2022.10068703}}


\bibitem[Wu et~al\mbox{.}(2021)]%
        {10.1145/3442381.3450015}
\bibfield{author}{\bibinfo{person}{Le Wu}, \bibinfo{person}{Lei Chen}, \bibinfo{person}{Pengyang Shao}, \bibinfo{person}{Richang Hong}, \bibinfo{person}{Xiting Wang}, {and} \bibinfo{person}{Meng Wang}.} \bibinfo{year}{2021}\natexlab{}.
\newblock \showarticletitle{Learning Fair Representations for Recommendation: A Graph-based Perspective}. In \bibinfo{booktitle}{\emph{Proceedings of the Web Conference 2021}} (Ljubljana, Slovenia) \emph{(\bibinfo{series}{WWW '21})}. \bibinfo{publisher}{Association for Computing Machinery}, \bibinfo{address}{New York, NY, USA}, \bibinfo{pages}{2198–2208}.
\newblock
\showISBNx{9781450383127}
\href{https://doi.org/10.1145/3442381.3450015}{doi:\nolinkurl{10.1145/3442381.3450015}}


\bibitem[Xia et~al\mbox{.}(2019)]%
        {XIA2019104857}
\bibfield{author}{\bibinfo{person}{Bin Xia}, \bibinfo{person}{Junjie Yin}, \bibinfo{person}{Jian Xu}, {and} \bibinfo{person}{Yun Li}.} \bibinfo{year}{2019}\natexlab{}.
\newblock \showarticletitle{WE-Rec: A fairness-aware reciprocal recommendation based on Walrasian equilibrium}.
\newblock \bibinfo{journal}{\emph{Knowledge-Based Systems}}  \bibinfo{volume}{182} (\bibinfo{year}{2019}), \bibinfo{pages}{104857}.
\newblock
\showISSN{0950-7051}
\href{https://doi.org/10.1016/j.knosys.2019.07.028}{doi:\nolinkurl{10.1016/j.knosys.2019.07.028}}


\bibitem[Yang et~al\mbox{.}(2023)]%
        {10.1145/3604915.3608784}
\bibfield{author}{\bibinfo{person}{Hao Yang}, \bibinfo{person}{Zhining Liu}, \bibinfo{person}{Zeyu Zhang}, \bibinfo{person}{Chenyi Zhuang}, {and} \bibinfo{person}{Xu Chen}.} \bibinfo{year}{2023}\natexlab{}.
\newblock \showarticletitle{Towards Robust Fairness-aware Recommendation}. In \bibinfo{booktitle}{\emph{Proceedings of the 17th ACM Conference on Recommender Systems}} (Singapore, Singapore) \emph{(\bibinfo{series}{RecSys '23})}. \bibinfo{publisher}{Association for Computing Machinery}, \bibinfo{address}{New York, NY, USA}, \bibinfo{pages}{211–222}.
\newblock
\showISBNx{9798400702419}
\href{https://doi.org/10.1145/3604915.3608784}{doi:\nolinkurl{10.1145/3604915.3608784}}


\bibitem[Yang et~al\mbox{.}(2020)]%
        {yang2020causalintersectionalityfairranking}
\bibfield{author}{\bibinfo{person}{Ke Yang}, \bibinfo{person}{Joshua~R. Loftus}, {and} \bibinfo{person}{Julia Stoyanovich}.} \bibinfo{year}{2020}\natexlab{}.
\newblock \bibinfo{title}{Causal intersectionality for fair ranking}.
\newblock
\showeprint[arxiv]{2006.08688}~[cs.LG]
\urldef\tempurl%
\url{https://arxiv.org/abs/2006.08688}
\showURL{%
\tempurl}


\bibitem[Yao and Huang(2017)]%
        {NIPS2017_e6384711}
\bibfield{author}{\bibinfo{person}{Sirui Yao} {and} \bibinfo{person}{Bert Huang}.} \bibinfo{year}{2017}\natexlab{}.
\newblock \showarticletitle{Beyond Parity: Fairness Objectives for Collaborative Filtering}. In \bibinfo{booktitle}{\emph{Advances in Neural Information Processing Systems}}, \bibfield{editor}{\bibinfo{person}{I.~Guyon}, \bibinfo{person}{U.~Von Luxburg}, \bibinfo{person}{S.~Bengio}, \bibinfo{person}{H.~Wallach}, \bibinfo{person}{R.~Fergus}, \bibinfo{person}{S.~Vishwanathan}, {and} \bibinfo{person}{R.~Garnett}} (Eds.), Vol.~\bibinfo{volume}{30}. \bibinfo{publisher}{Curran Associates, Inc.}, \bibinfo{address}{Red Hook, NY, USA}.
\newblock
\urldef\tempurl%
\url{https://proceedings.neurips.cc/paper_files/paper/2017/file/e6384711491713d29bc63fc5eeb5ba4f-Paper.pdf}
\showURL{%
\tempurl}


\bibitem[Zafar et~al\mbox{.}(2017)]%
        {pmlr-v54-zafar17a}
\bibfield{author}{\bibinfo{person}{Muhammad~Bilal Zafar}, \bibinfo{person}{Isabel Valera}, \bibinfo{person}{Manuel~Gomez Rogriguez}, {and} \bibinfo{person}{Krishna~P. Gummadi}.} \bibinfo{year}{2017}\natexlab{}.
\newblock \showarticletitle{{Fairness Constraints: Mechanisms for Fair Classification}}. In \bibinfo{booktitle}{\emph{Proceedings of the 20th International Conference on Artificial Intelligence and Statistics}} \emph{(\bibinfo{series}{Proceedings of Machine Learning Research}, Vol.~\bibinfo{volume}{54})}, \bibfield{editor}{\bibinfo{person}{Aarti Singh} {and} \bibinfo{person}{Jerry Zhu}} (Eds.). \bibinfo{publisher}{PMLR}, \bibinfo{pages}{962--970}.
\newblock
\urldef\tempurl%
\url{https://proceedings.mlr.press/v54/zafar17a.html}
\showURL{%
\tempurl}


\bibitem[Zehlike et~al\mbox{.}(2017)]%
        {10.1145/3132847.3132938}
\bibfield{author}{\bibinfo{person}{Meike Zehlike}, \bibinfo{person}{Francesco Bonchi}, \bibinfo{person}{Carlos Castillo}, \bibinfo{person}{Sara Hajian}, \bibinfo{person}{Mohamed Megahed}, {and} \bibinfo{person}{Ricardo Baeza-Yates}.} \bibinfo{year}{2017}\natexlab{}.
\newblock \showarticletitle{FA*IR: A Fair Top-k Ranking Algorithm}. In \bibinfo{booktitle}{\emph{Proceedings of the 2017 ACM on Conference on Information and Knowledge Management}} (Singapore) \emph{(\bibinfo{series}{CIKM '17})}. \bibinfo{publisher}{Association for Computing Machinery}, \bibinfo{address}{New York, NY, USA}, \bibinfo{pages}{1569–1578}.
\newblock
\showISBNx{9781450349185}
\href{https://doi.org/10.1145/3132847.3132938}{doi:\nolinkurl{10.1145/3132847.3132938}}


\bibitem[Zhu et~al\mbox{.}(2018)]%
        {zhu18}
\bibfield{author}{\bibinfo{person}{Ziwei Zhu}, \bibinfo{person}{Xia Hu}, {and} \bibinfo{person}{James Caverlee}.} \bibinfo{year}{2018}\natexlab{}.
\newblock \showarticletitle{Fairness-Aware Tensor-Based Recommendation}. In \bibinfo{booktitle}{\emph{Proceedings of the 27th ACM International Conference on Information and Knowledge Management}}. \bibinfo{publisher}{Association for Computing Machinery}, \bibinfo{address}{New York, NY, USA}, \bibinfo{pages}{1153--1162}.
\newblock
\href{https://doi.org/10.1145/3269206.3271795}{doi:\nolinkurl{10.1145/3269206.3271795}}


\bibitem[Zhu et~al\mbox{.}(2020)]%
        {10.1145/3397271.3401177}
\bibfield{author}{\bibinfo{person}{Ziwei Zhu}, \bibinfo{person}{Jianling Wang}, {and} \bibinfo{person}{James Caverlee}.} \bibinfo{year}{2020}\natexlab{}.
\newblock \showarticletitle{Measuring and Mitigating Item Under-Recommendation Bias in Personalized Ranking Systems}. In \bibinfo{booktitle}{\emph{Proceedings of the 43rd International ACM SIGIR Conference on Research and Development in Information Retrieval}} (Virtual Event, China) \emph{(\bibinfo{series}{SIGIR '20})}. \bibinfo{publisher}{Association for Computing Machinery}, \bibinfo{address}{New York, NY, USA}, \bibinfo{pages}{449–458}.
\newblock
\showISBNx{9781450380164}
\href{https://doi.org/10.1145/3397271.3401177}{doi:\nolinkurl{10.1145/3397271.3401177}}


\end{thebibliography}

\appendix
\section{Appendix}



\subsection{Abalation Study}\label{sec:abalation}
\begin{figure}[h]
    \centering
    \includegraphics[width=0.5\textwidth]{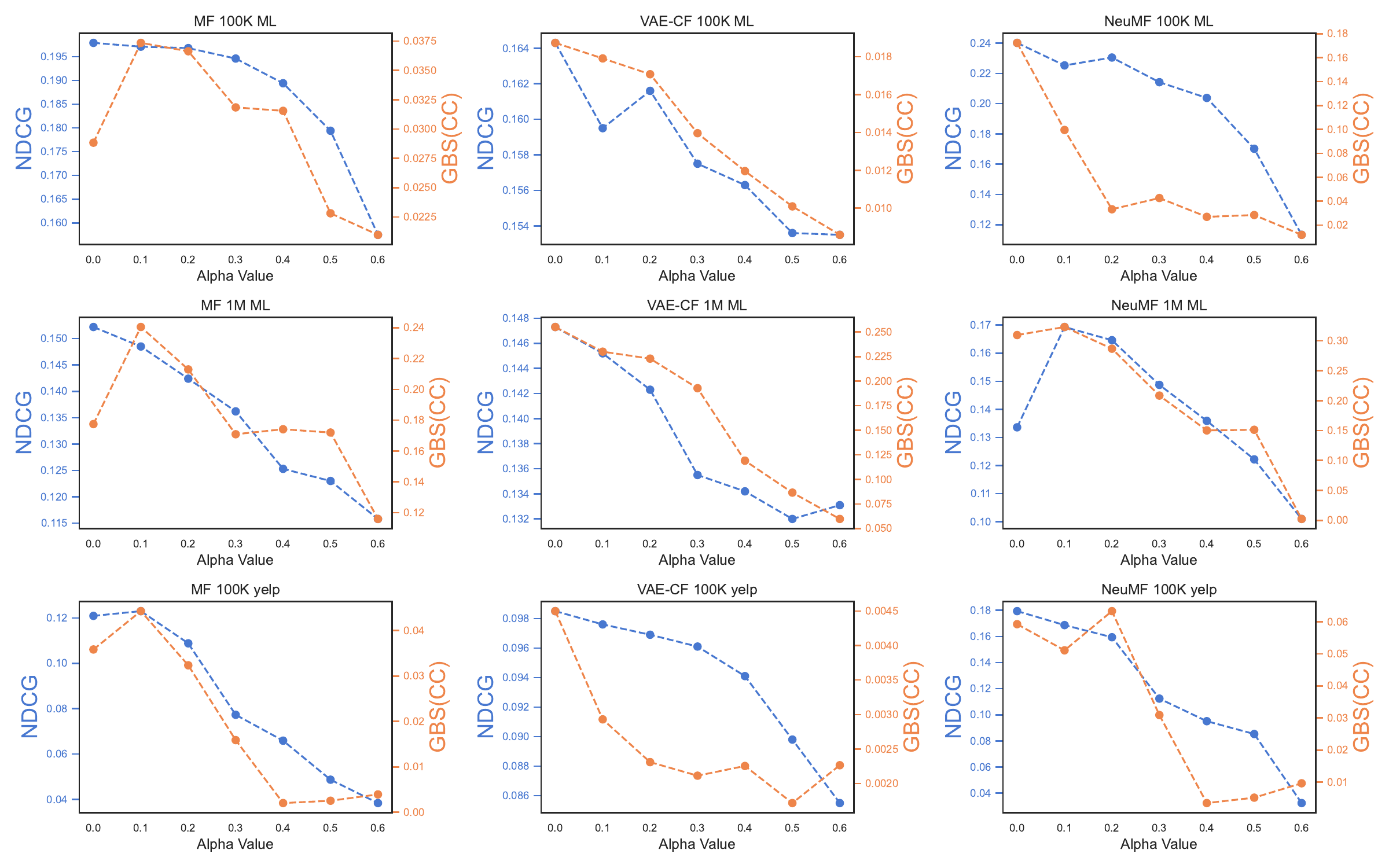}
    \caption{Impact of \(\alpha\) on recommendation performance wrt NDCG@50 and the bias measure which is the difference of Category Coverage values for male and female. The experiments are performed for all three  datasets }
    \label{fig:100kabalation}
\end{figure}

In this section, we discuss the influence of the hyper-parameter \(\alpha\) on recommendation performance and fairness. As mentioned in Section \ref{sec:fairloss}, \(\alpha\) can be used to control the strength of the fairness aware loss. Theoretically as \(\alpha\) increases we would expect a decrease in performance since the recommendation loss would have a decreased weight. We use \(\alpha\) values from 0 to 0.6 with increments of 0.1 for all three models. We did not include values above 0.6, because it does not make sense to overpower the recommendation loss since that is our main task. To verify the expected behavior of alpha on recommendation loss and bias we plot the metric we optimize on: GBS(CC) or \(\bigtriangleup Category Coverage\) and the NDCG values. As shown in Figure \ref{fig:100kabalation}, there is a general trend of the performance metric and the bias score dropping as \(\alpha\) increases. There are some fluctuations, where the bias increases for the first few values of alpha. Our intuition for this is that the fairness loss might not be enough to decrease the bias when \(\alpha\) is small. It can essentially end up disturbing the loss function itself, without explicitly decreasing bias. But as the value increases the bias scores drop lower which is expected. There is a slight increase in NDCG value for certain models, as mentioned in Section \ref{sec:perform-evaluation}, this can be the result of the fairness loss term acting as a regularization term that essentially prevents over-fitting in the model. We choose \(\alpha\) values by ensuring there is a decrease in bias, without significant loss in performance. For the ML 100K dataset, we choose 0.4, 0.3, and 0.2 respectively for MF, VAE-CF, and NeuMF models. For the ML 1M dataset, we choose 0.6,0.2, and 0.4 for MF, VAE-CF, and NeuMF models respectively. Lastly, for the Yelp dataset, we choose 0.3 for all models. 

\subsection{Details about Baseline Models}\label{sec:baseline}
This subsection delves into details about the models we worked on for our experiments.
\begin{itemize}[left=5pt]
    \item MF \cite{5197422}: A classical Matrix Factorization algorithm where users and items are represented as latent vectors with global bias. 
    In our implementation, we use BPR \cite{10.5555/1795114.1795167} loss with negative sampling to enhance the performance of the model.
    \item UserKNN \cite{10.5555/2074094.2074100}:  This is a neighborhood-based method based on users, and items are recommended by discovering similar users based on the cosine similarity of their historical interactions. 
    \item ItemKNN \cite{10.1145/352871.352887}: Another neighborhood-based method that computes the similarity between items instead. 
    \item NeuMF \cite{10.1145/3038912.3052569}: NeuMF is a deep-learning based extension to MF, which combines the linearity of traditional MF models and the non-linearity of DNNs (Deep Neural Networks).
    \item VAE-CF \cite{10.1145/3178876.3186150}: This non-linear probabilistic model is based on the auto-encoder architecture and learns compressed information about data. The encoder helps map user interactions as a low-dimensional latent space, and the decoder decodes this information back to the high-dimensional vector which is used to make predictions.
\end{itemize}

The \(U_{val}\) regularization term for BeyondParity \cite{NIPS2017_e6384711} can be written as:
\begin{equation}
    U_{\text{val}} = \frac{1}{n} \sum_{j=1}^{n} \left| \left( E_{g} [y]_{j} - E_{g} [r]_{j} \right) - \left( E_{\neg g} [y]_{j} - E_{\neg g} [r]_{j} \right) \right|
    \label{eq:bp}
\end{equation}

where \( E_{g} [y]_{j} \) is the average predicted score for the \( j \)-th item from disadvantaged users, \( E_{\neg g} [y]_{j} \) is the average predicted score for the \( j \)-th item from advantaged users, \( E_{g} [r]_{j} \) is the average rating for the \( j \)-th item from disadvantaged users and \( E_{\neg g} [r]_{j} \) is the average rating for the \( j \)-th item from advantaged users. This fairness objective is optimized alongside the actual learning objective for a collaborative-based recommendation system. SM-GBiasedMF, is a fair recommendation model which achieves counterfactual fairness by utilizing adversarial learning. They combine recommendation loss and an adversarial loss and, the trade-off is controlled by \(\lambda\).
\subsection{Dataset Pre-processing}
For all of our datasets, we filter out inactive users. A user is considered inactive if they have less than 5 interactions. Additionally, we remove any user whose gender is not known. The Yelp dataset has over 300 categories. We reduce this set of categories to a condensed group of 21 broader categories, for better interpretability. These categories include \textit{Active Life \& Fitness}, \textit{Arts \& Entertainment}, \textit{Automotive, Bars \& Nightlife}, \textit{Coffee, Tea \& Desserts}, \textit{Drinks \& Spirits, Education \& Learning}, \textit{Event Services, Family \& Kids}, \textit{Food \& Restaurants}, \textit{Health \& Beauty}, \textit{Home \& Garden}, \textit{Miscellaneous}, \textit{Outdoor Activities}, \textit{Public Services \& Community}, \textit{Shopping \& Fashion},\textit{ Specialty Food \& Groceries}, \textit{Sports \& Recreation},\textit{ Technology \& Electronics}, \textit{Travel \& Transportation}, and \textit{Asian}.

\subsection{Process of generating Figure \ref{fig:compare_genre}}
This plot is generated using the ML-100K dataset. Once the models are done being trained we identify the users by their gender in the test set. To ensure a fair comparison we take the number of users in the smaller group, which was female in our case. We randomly chose the same number of male users from the test set. Once we have an equal number of male and female users, precision@10 values are calculated for each user and averaged by gender. To find the proportions of movies recommended, we generate the top 10 movies for each user. The proportion of each genre is computed using Equation \ref{eq:M1}.

\end{document}